\documentclass[lettersize,journal]{IEEEtran}
\usepackage{amssymb,amsmath,amsfonts}
\usepackage{algorithmic}
\usepackage{algorithm}
\usepackage{array}
\usepackage{booktabs}
\usepackage{colortbl}
\usepackage[hidelinks]{hyperref}
\usepackage{graphicx}
\usepackage{makecell}
\usepackage[mathlines, switch]{lineno}
\usepackage{multirow}
\usepackage[sort&compress, numbers]{natbib}
\usepackage{pifont}
\usepackage{stfloats}
\usepackage[caption=false,font=footnotesize]{subfig}
\usepackage{textcomp}
\usepackage{url}
\usepackage{verbatim}
\usepackage[dvipsnames]{xcolor}
\usepackage{subcaption}
\usepackage{tabularx}
\usepackage{placeins}

\hypersetup{
  colorlinks = true,
  urlcolor = CadetBlue,
  linkcolor = Cerulean,
  citecolor = Maroon
}

\definecolor{whitesmoke}{rgb}{0.96, 0.96, 0.96}

\begin{document}

\title{HOPE: A Reinforcement Learning-based Hybrid Policy Path Planner for Diverse Parking Scenarios}

\author{Mingyang Jiang, Yueyuan~Li, Songan~Zhang, Siyuan~Chen, Chunxiang~Wang, and~Ming~Yang
\thanks{This work is supported in part by the National Natural Science Foundation of China under Grant 62173228, Grant 62373250, Grant U22A20100, and Grant 52402504 \textit{(Corresponding author: Ming Yang, email: MingYANG@sjtu.edu.cn)}.}
\thanks{Mingyang Jiang, Yueyuan Li, Siyuan Chen, Chunxiang Wang, and Ming Yang are with the Department of Automation, Shanghai Jiao Tong University, Key Laboratory of System Control and Information Processing, Ministry of Education of China, Shanghai, 200240, CN (Mingyang Jiang and Yueyuan Li are co-first authors).}
\thanks{Songan Zhang is with the Global Institute of Future Technology, Shanghai Jiao Tong University, Shanghai, 200240, CN}
}



\maketitle

\begin{abstract}
Automated parking stands as a highly anticipated application of autonomous driving technology. However, existing path planning methodologies fall short of addressing this need due to their incapability to handle the diverse and complex parking scenarios in reality. While non-learning methods provide reliable planning results, they are vulnerable to intricate occasions, whereas learning-based ones are good at exploration but unstable in converging to feasible solutions. To leverage the strengths of both approaches, we introduce Hybrid pOlicy Path plannEr (HOPE). This novel solution integrates a reinforcement learning agent with Reeds-Shepp curves, enabling effective planning across diverse scenarios. HOPE guides the exploration of the reinforcement learning agent by applying an action mask mechanism and employs a transformer to integrate the perceived environmental information with the mask. To facilitate the training and evaluation of the proposed planner, we propose a criterion for categorizing the difficulty level of parking scenarios based on space and obstacle distribution. Experimental results demonstrate that our approach outperforms typical rule-based algorithms and traditional reinforcement learning methods, showing higher planning success rates and generalization across various scenarios. 
We also conduct real-world experiments to verify the practicability of HOPE.
The code for our solution is openly available on \href{GitHub}{https://github.com/jiamiya/HOPE}.
\end{abstract}


\section{Introduction}
 Automated parking is a tempting technology to improve driving safety and efficiency \cite{song2016analysis}. 
 An automated parking system comprises several vital components, including perception, planning, and control, in which path-planning algorithms play a crucial role \cite{wang2014automatic}. 
In parking scenarios, the path-planning task involves generating a feasible path from the start position to the target parking spots under specific physics constraints.
Compared to other scenarios, finding a feasible path in parking scenarios is usually more challenging because of the lower error tolerance of the target spot and the lack of navigational reference lines \cite{li2021optimization}. Additionally, the limited space and surrounding obstacles shrink the number of potential solutions.
While existing path-planning approaches have proved practical in most simple scenarios, their inherent difficulty in understanding the surrounding environment may lead to planning failures, especially when scenarios become more intricate \cite{likmeta2020combining}.


Learning-based planners have the potential to understand environments and intelligently plan routes through a data-driven approach, diverging from reliance on pre-defined planning methods rooted in human priors. While expert data is used as the ground truth for imitation learning in various tasks, the scarcity of large-scale datasets for parking scenarios necessitates researchers to collect data manually \cite{teng2023motion}. This supervised learning approach risks overfitting the model to specific parking strategies due to the insufficient diversity in the training scenarios.
Meanwhile, Reinforcement Learning (RL)  has gathered increasing attention in the field of autonomous driving \cite{kiran2021deep}. Indeed, through interaction with the environment, RL methods enable the training of agents without labeling trajectory ground truth. Nonetheless, training RL agents in complex and diverse scenarios remains a challenging task \cite{grigorescu2020survey}.
It is straightforward for the agent to fit into fixed parking strategies but poses greater difficulty in obtaining generalization capability across various parking scenarios.
Moreover, the agent faces challenges in exploring effectively, especially in complex scenarios with narrow parking spaces, significantly impacting the training efficiency.

This paper focuses on employing RL methods in the parking path planning task with static obstacles.
To achieve efficient and effective learning under diverse parking scenarios, we propose reinforcement learning-based \textbf{H}ybrid p\textbf{O}licy \textbf{P}ath plann\textbf{E}r (HOPE). The hybrid policy planner is designed to leverage RL-based methods and a classical geometric-based path planning method, the Reeds-Shepp (RS) curve \cite{reeds1990optimal}. 
A transformer-based structure is used as the information fusion network in actor and critic networks \cite{vaswani2017attention}.
Since the diversity of scenario difficulty has a significant impact on training and testing, we rank the difficulty of scenarios by referencing related standards for automated parking.
Overall, the main contributions of this paper include:
\begin{itemize}
    \item We develop a hybrid policy method for parking path planning which achieves over 97\% success rate across diverse and challenging parking scenarios and verifies its generalization ability with real-world experiments.
    \item We propose to implement an action mask in the path planning task. This mechanism excludes improper actions for reinforcement learning agents and significantly improves training efficiency and performance.
    \item We propose a criterion to rank the difficulty of static parking scenarios. Comprehensive experiments are conducted across simulations of different difficulty levels, demonstrating a notable improvement in success rates compared to rule-based and naive reinforcement learning-based methods.
\end{itemize}

\section{Related Work}

\subsection{Non-learning-based parking path-planning}
Non-learning-based path planning methods under parking scenarios mainly include geometric-based methods and sampling-and-search-based methods \cite{li2021optimization}. Geometric planners construct paths connecting the start position and the target position using different kinds of geometric curves \cite{liang2012automatic,vorobieva2014automatic2}. 
Typical geometric methods include the Dubins curves and the RS curves \cite{dubins1957curves,reeds1990optimal}. Building upon the RS curve, several improvements have been made to enhance its performance in parking scenarios \cite{du2014autonomous,kim2014auto}.
The sampling-and-search-based approach discretizes and searches either in the state space or in the control space to find a valid path. 
One widely employed approach is the Hybrid A* method \cite{dolgov2010path}, which devises a variant of the A* algorithm that incorporates the kinematic state space of the vehicle to derive a kinematically feasible trajectory. Hybrid A* was initially utilized in the Defense Advanced Research Projects Agency Challenge (DARPA) and has since undergone improvements and application to path planning tasks in parking scenarios \cite{sedighi2019guided,sheng2021autonomous}.
Both geometric and sampling-and-search-based methods, as rule-based methods, leverage human prior knowledge to design algorithms. In most common cases, these priors can serve as a strong fallback to obtain satisfactory solutions. 
However, these methods can hardly achieve human-level proficiency in complex and diverse parking scenarios \cite{czubenko2015autonomous}.

Another category of methods involving path planning is the optimization-based trajectory planning approach.
 These methods solve the trajectory planning problem by formulating it as an optimal control problem.
 Related works have made improvements in collision avoidance constraint formulation \cite{zhang2020optimization}, iterative solution efficiency \cite{li2021optimization}, and robustness \cite{he2021tdr}. 
While this approach can produce smoother paths adhering to vehicle kinematics, the optimization process relies on another path-planning algorithm, often the Hybrid A* method \cite{han2023efficient, zhang2020optimization, he2021tdr}, to obtain initial solutions. This reliance helps improve computational efficiency and allows these methods to focus on optimizing smoother trajectories, as well as determining the limit of planning success rate.
 

\subsection{Learning-based parking path-planning}
Learning-based methods represent a potential avenue to enhance the planner's performance in the task of parking. Existing methods mainly include imitating learning-based and reinforcement learning-based methods. Imitating learning requires the learnable planner to fit on the ground truth data. Liu et al. employed a neural network to align parking paths more closely with human behavior \cite{liu2017parking}. Rathour S et al. derived a parking policy by imitating trajectories collected from expert drivers \cite{rathour2018vision}. 
Other works include implementing the deep neural network or deep recurrent neural network as the behavior cloner on real vehicles \cite{chai2020design,chai2022deep}.
However, imitation learning-based methods are not good at handling the scenarios beyond their training sets, and the behavior cloners cannot outperform their imitation targets in terms of performance \cite{ho2016generative}.

In RL-based approaches, the agents are developed without any labeled ground truth.
The Monte Carlo tree search method was applied to search the available path in parallel parking scenarios \cite{song2022time}. Bernhard J et al. learned the heuristics with Deep Q-learning (DQL) and improved the path search process in the Hybrid A* algorithm \cite{mnih2015human,bernhard2018experience}. Du et al. directly employed DQL to train an agent to produce single-step path planning results iteratively in a parallel and vertical case \cite{du2020trajectory}. 
Yuan et al. proposed a hierarchical planning approach where high-level RL agents are trained to generate initial reference solutions \cite{yuan2023hierarchical}.
The existing works typically map the elements of the task to the components of RL and directly employ them to train the agent in a limited number of scenarios. As more complex and diverse parking scenarios are introduced, it would be harder for RL-based methods to explore the proper parking policy, making it challenging to obtain planners with generalization capabilities.

\section{Preliminaries}

\subsection{Reinforcement learning for path planning}
The RL problem can be addressed as policy learning in a Markov decision process (MDP) defined by a 4-element tuple $(\mathcal{S}, \mathcal{A}, p, r)$, where $\mathcal{S}$ is the state space, and $\mathcal{A}$ is the action space. In the path-planning problem, $s_t \in \mathcal{S}$ includes the vehicle position and orientation $p_t=(x_t, y_t, \theta_t)$ as well as other observable information about obstacles and target parking space.
We selected velocity $v$ and steering angle $\delta$ as the action space $\mathcal A=\{ a = (v, \delta) \}$.
$p$ denotes the state transition probability density of the next state $s_{t+1}$ given the current state $s_t$ and action $a_t$. In practice, the transition is modeled using the single-track bicycle model, as implemented in \cite{althoff2017commonroad}, and this uncertainty is not considered in a deterministic environment. A reward $r=r(s_{t+1}, a_t)$ is given by the environment $E$ based on the state and action in each interaction step.
The objective is to learn the policy $\pi(a_t|s_t)$ that maximizes the future reward  $R_t = \sum_{i=t}^{T} \gamma^{(i-t)} r(s_i, a_i)$  with a discounting factor $\gamma \in [0,1]$. Given the start point $p_0$ and the target $p_T$, 
the feasible parking path \(P = \{p_0, p_1, p_2, ..., p_T\}\) can be obtained iteratively using the policy $\pi$.


To obtain the optimal policy, reinforcement learning algorithms primarily use the action-value function to describe the reward in terms of expectation after taking action $a_t$ in state $s_t$ with policy $\pi$:
\begin{equation}
\begin{aligned}
    Q_\pi(s_t, a_t) &= E_{i>t, s_i \sim E,a_i \sim \pi} [R_t|s_t, a_t] \\
    V_\pi(s_t) &= E_{a_t \sim \pi(\cdot|s_t)}[Q_\pi(s_t, a_t)],
\end{aligned}
\end{equation}
and both the Q function and value function can be updated by the Bellman equation:
\begin{equation}
\begin{aligned}
Q_\pi(s_t, a_t) &= E_{r_t, s_{t+1} \sim E} [ r(s_t, a_t) + \gamma Q_\pi(s_{t+1}, a_{t+1}) ] \\
V_\pi(s_t) &= E_{a_t \sim \pi(\cdot|s_t)}[r(s_t, a_t) + \gamma V_\pi(s_{t+1})].
\end{aligned}
\label{eq:bellman}
\end{equation}

To demonstrate the improvement of our proposed method across different reinforcement learning approaches, in this paper, we chose the commonly used on-policy algorithm, Proximal Policy Gradient (PPO) \cite{schulman2017proximal}, and off-policy algorithm, Soft Actor-Critic (SAC) \cite{haarnoja2018soft}, to obtain the reinforcement learning policy $\pi_\theta$ in HOPE.

\subsubsection{PPO}
PPO is a policy-gradient method that performs the gradient ascent in the trust region. To model the change between the old policy parameters $\pi_{\theta_{\text{old}}}$ before the update and the new parameters $\pi_{\theta}$, a probability ratio is denoted as $r_t(\theta) = \frac{\pi_\theta(a_t|s_t)}{\pi_{\theta_{\text{old}}}(a_t|s_t)}$. The loss function is:
\begin{equation}
    L(\theta) = \hat{E}_t \left[ \min\left(r_t(\theta) \hat{A}_t, \text{clip}(r_t(\theta), 1 - \epsilon, 1 + \epsilon) \hat{A}_t \right) \right]
    \label{eq: PPO loss}
\end{equation}
where $\hat{A}_t=Q(s_t,a_t) - V(s_t)$ is the estimated advantage value at time step $t$. $\epsilon$ is a hyperparameter indicating how much $r_t(\theta)$ can deviate from $1$. 
By taking the minimum of the two terms, 
the policy is updated within a certain range in each iteration to achieve stable and efficient training.

\subsubsection{SAC}
SAC is a maximum entropy reinforcement learning method that maximizes the cumulated reward as well as policy entropy. The loss function of SAC can be expressed as:
\begin{equation}
    L(\theta) = E_{s_t \sim D} \left[ \alpha \mathcal{H}(\pi(\cdot | s_t)) - Q(s_t, a_t) \right].
    \label{eq: SAC loss}
\end{equation}
$\mathcal{H}(\pi(\cdot | s_t) = \log \pi_\theta(a_t | s_t)$ is the augmented entropy term and is scaled with a temperature parameter $\alpha$. With the entropy regularization, SAC encourages the agent to keep a diverse policy and explore different strategies.

\subsection{Reeds-Shepp curve}
RS curve is designed to generate the shortest path that links two positions with the minimal steering radius of the vehicle and straight lines. It is mathematically proved that the shortest path belongs to 48 types of curves and can be represented by one of the following 9 expressions \footnote{Visualization examples of RS curves can be found in \url{https://tactics2d.readthedocs.io/en/latest/tutorial/interpolator_visualization/}}:
\begin{equation}
\begin{aligned}
    & C|C|C,\ CC|C,\ CSC,\ CC_u|C_uC,\ C|C_uC_u|C,\\
    & C|C_{\pi/2}SC,\ C|C_{\pi/2}SC_{\pi/2}|C,\ C|C|C,\ CSC_{\pi/2}|C .
\end{aligned}
\label{eq: rs}
\end{equation}

Here, $C$ represents an arc segment with left or right steering, and $S$ represents a straight-line segment. $|$ means the path segment after it uses an opposite direction compared to the one before it. $C_{\pi/2}$ refers to a circular arc in which the central angle is fixed at $\pi/2$, and the arcs noted $C_u$ in one formula share the same central angle $u$.

When calculating the RS curve, the path scale is initially normalized based on the vehicle's minimal turning radius $r_{min}$. Subsequently, the lengths of all segments for each type of path are determined using the given scaled starting point $p_0 = (x_0/r_{min}, y_0/r_{min}, \theta_0)$, ending point $p_T = (x_T/r_{min}, y_T/r_{min}, \theta_T)$, and geometric constraints specified by the expression \ref{eq: rs}. The total length of each path is obtained by summing the lengths of all segments, and the curve with the shortest total length is then selected as the optimal RS curve. 
The path waypoints $P= \{p_0, p_1, ..., p_T\}$ along the entire curve can be obtained through interpolation based on the lengths and types of each curve segment. Additionally, the vehicle's steering angle at each waypoint can be calculated using the vehicle model.

\section{Methodology}
\begin{figure*}
  \centering
  \includegraphics[width=\textwidth]{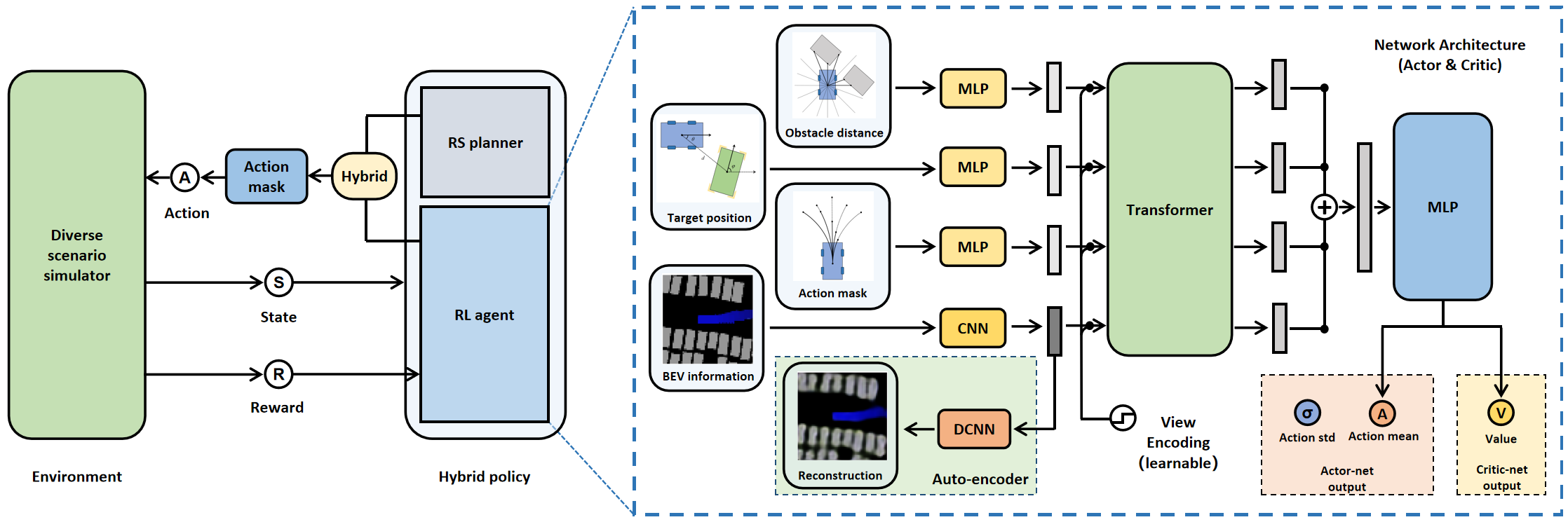}
  \caption{The overall structure of the proposed method, including the interaction loop with the simulator (left) and the network architecture (right).}
  \label{fig:overall_structure}
\end{figure*}

\subsection{Overview of architecture}
The overall RL-based path planning framework and network structure are shown in Figure \ref{fig:overall_structure}. 
To enhance training efficiency, we employ a hybrid policy reinforcement learning approach. Our proposed HOPE combines the learnable policy from original reinforcement learning with a rule-based policy derived from the RS curve.
In each interaction step, the agent outputs the action, namely the single-step path planning result, based on the current state given by the environment. The action is then adjusted using the action mask mechanism before interacting with the environment.
We employ four forms of network input as state representation, including:
\begin{itemize}
    \item Obstacle distance (vector-based) $l_t$: the value of the nearest distance to obstacles at certain angles, which will be introduced later in section \ref{sec: action mask}.
    \item Target position (vector-based) $P_{tgt}$: $P_{tgt}$ is defined as a 5-elements tuple $(d, \text{cos}(\theta_t), \text{sin}(\theta_t), \text{cos}(\phi_t), \text{sin}(\phi_t))$, includes the information about distance to target parking spot's position $d$, orientation $\theta_t$, and heading $\phi_t$ in the ego vehicle's coordinate system.
    \item Action mask (vector-based) $f_{am}$: the representation of max valid step size at different steering angles, which will be introduced in section \ref{sec: action mask 2}.
    \item Bird-eye-view information (image-based) $I_{BEV}$: a low-resolution depiction of the drivable area, target parking spot, and the historical trajectory of the ego vehicle. 
\end{itemize}
A transformer-based structure with learnable view encoding is designed to fuse the inputs and get the outputs in the actor and critic network \cite{shao2023safety}. We also utilize the auto-encoder structure to pre-train the image encoder.
More details of our network and reward function are discussed in the appendix \ref{appendix: state} and \ref{appendix: reward}.
The pseudo-code of our method is shown in Algorithm \ref{algo:algorithm}.

\vspace{-5pt}
\begin{algorithm}[thb]
  \caption{The training process of HOPE}
  \label{algo:algorithm}
\begin{algorithmic}
  \STATE Initialize the simulation environment E, the agent policy $\pi_\theta$, the value network $V_\psi\text{ and the Q network } Q_\phi$, the hybrid policy $\pi_h$, the replay buffer $D$
  \STATE \textbf{for} iteration $k = 0,1,2,\dots$ \textbf{do}
  \STATE \hspace{0.4cm} Reset the environment: $t \leftarrow 0$,  $s_t, terminate \leftarrow$ E
  \STATE \hspace{0.4cm} \textbf{while} $terminate$ is False:
  \STATE \hspace{0.8cm} Choose action with the hybrid policy $a_t = \pi_h(s_t)$
  \STATE \hspace{0.8cm} Get the masked action $\hat{a}_t$ using Equation \ref{eq: action mask 1}, \ref{eq: action mask 2}
  \STATE \hspace{0.8cm} Interact with E: $s_{t+1}, r_t, terminate = E(\hat{a}_t)$
  \STATE \hspace{0.8cm} Add to $D$: $D = D \cup (s_t, \hat{a}_t, r_t, s_{t+1}, terminate)$
  \STATE \hspace{0.8cm} \textbf{if} update condition reached \textbf{then}
  \STATE \hspace{1.2cm} (PPO) Update $\theta, \psi$ using Equation. \ref{eq: PPO loss}, \ref{eq: new bellman}
  \STATE \hspace{1.2cm} (SAC) Update $\theta, \phi$ using Equation. \ref{eq: new bellman}, \ref{eq: new SAC loss}
  \STATE \hspace{0.8cm} \textbf{end if}
  \STATE \hspace{0.8cm} $t \leftarrow t+1$
  \STATE \hspace{0.4cm} \textbf{end while}
  \STATE \textbf{end for}
\end{algorithmic}
\end{algorithm}
\vspace{-10pt}

\subsection{Hybrid policy}
To improve the exploring and training efficiency, we combine the RL policy $\pi_\theta$ and the derived RS policy $\pi_{RS}$ to facilitate the planning process. The hybrid process at timestamp $t$ can be expressed as a function $h_t: \mathcal{A} \times \mathcal{A} \rightarrow \mathcal{A}$ that maps the actions $a_t \sim \pi_\theta(\cdot|s_t)$ from RL policy and $a'_t \sim \pi_{RS}(\cdot|s_t)$ from RS policy to the hybridized action $\tilde{a}_t = h_t(a_t, a'_t) \sim \pi_h(\cdot|s_t)$. 
Such an action hybrid process can be regarded as the procedure of making a suitable switch between two actions. Specifically, it activates the rule-based policy at certain timestamps and otherwise the reinforcement learning policy is used.

\subsubsection{Rule-based policy from Reeds-Shepp curve}
The nine expressions in Equation \ref{eq: rs} represent 48 types of curves that give the shortest path in free space. However, in the presence of obstacles, the shortest RS curve may not be feasible. 
To implement the RS method in the parking scenario, we make two modifications in practice:
\begin{itemize} 
\item We calculate the shortest K curves and assess their feasibility in ascending order of length, selecting the shortest feasible, collision-free path as the final route.
\item We include paths of the $S(|)C(|)S$ form. These paths are excluded in the original RS method because they are strictly suboptimal in obstacle-free spaces. However, in environments with obstacles, these paths may still be feasible and beneficial, as indicated in Appendix \ref{appendix: scs}.
\end{itemize}

\subsubsection{Exploring and learning with RS policy}

During training, the agent explores and interacts with the environment using the hybridized action $\tilde{a}_t$ rather than raw action $a_t$. This choice also affects the updates on the policy $\pi_\theta$ and Q function or value function. 
For the Bellman equation in Equation \ref{eq:bellman}, the update on the parameter $V_\psi$ and $Q_\phi$ can be written as:
\begin{equation}
\begin{aligned}
    J_V(\psi) &= E_{s_t \sim D_{\pi_h}} \left[ \frac{1}{2} \left( V_\psi(s_t) - E_{\tilde{a}_t \sim \pi_h} \left[ Q_\phi(s_t, \tilde{a}_t) \right] \right)^2 \right] \\
    J_Q(\phi) &=  \mathbb{E}_{(s_t, \tilde{a}_t) \sim D_{\pi_h}} 
     \left[ \frac{1}{2} \left( Q_\phi(s_t, \tilde{a}_t) \right. \right. \\
& \left. \left. - r(s_t, \tilde{a}_t) - \gamma \mathbb{E}_{s_{t+1} \sim \rho_{\pi_h}} \left[ V_\psi(s_{t+1}) \right] \right)^2 \right].
\end{aligned}
\label{eq: new bellman}
\end{equation}

Since the convergence of the Bellman equation for updating $V_\psi$ and $Q_\phi$ is independent of the specific policy, when we replace the policy $\pi$ with $\pi_h$, the iterative updates for $V_\psi$ and $Q_\phi$ still converge.
For policy updates in PPO, the probability ratio $r_t(\theta)$ in Equation \ref{eq: PPO loss} can be rewritten as:
\begin{equation}
    r_t(\theta) = \frac{\pi_\theta(a_t|s_t)}{\pi_{\theta_{old}}(\tilde{a}_t|s_t)}, \ \tilde{a}_t = h(a_t, a'_t).
    \label{eq: new ppo ratio}
\end{equation}
The ratio clipping operation prevents excessive gradient updates caused by a significant KL divergence between $\pi_\theta$ and $\pi_h$. 
Meanwhile, Equation \ref{eq: SAC loss} in SAC can be rewritten as:
\begin{equation}
    J_\pi(\theta) = E_{s_t \sim D_{\pi_h}} \left[ \log \pi_\theta(a_t | s_t) - Q_\phi(s_t, a_t) \right].
    \label{eq: new SAC loss}
\end{equation}

Generally, gradients on the Q function and value function are modified by using the data collected by $\pi_h$ instead of $\pi_\theta$, since they are now estimations on distribution $\rho_{\pi_h}$. 
We still update the original RL policy $\pi_\theta$ as shown in Equation \ref{eq: new ppo ratio} and \ref{eq: new SAC loss}, which indicates that the hybrid policy $\pi_h$, as well as the RL policy $\pi_\theta$, are optimized. 
In practice, we apply a switch strategy that the RS policy is activated only when: 1) the distance from the vehicle's position to the target position is smaller than a threshold $d_{rs}$, and 2) there exists a collision-free RS curve from the vehicle's position to the target position. 
By the hybrid policy approach, during the initial phases of the learning process when the agent is not well-trained, the RS method serves as an alternative strategy to offer additional positive examples for updates. The agent can thus learn how to adjust the vehicle's pose to explore feasible parking routes.

\subsection{Action mask}

Action masks have been employed in some discrete-space reinforcement learning tasks to filter out invalid actions, which can enhance training efficiency and ensure alignment with deployment conditions \cite{huang2020closer}. By incorporating action masks, agents can focus on making complex decisions without spending extensive time learning calculable constraints. Additionally, this approach helps prevent invalid behaviors in applications.
However, in path planning tasks, the computation of an action mask, which includes collision detection for all feasible actions of the agent vehicle across diverse state spaces, is computationally expensive.
Here, we present a method for calculating and utilizing the action mask to enhance the training efficiency of reinforcement learning in path-planning tasks.
Specifically, we use $Collide(s_t)$ to denote the event whether the vehicle at state $s_t$ collides with obstacles. The action mask provides information about the largest safe step velocity $v^*$ at any given steering angle $\delta$:
\begin{equation}
\begin{aligned}
    v^* &= \operatorname{max}_v \{Collide(s_{t+1}) = False\}, \\
\end{aligned}
\end{equation}
where $s_{t+1}$ is the new state after executing action $a=(v, \delta)$ at state $s_t$. By using $v^*$ to constraint the raw speed $v$ output by the actor net, we can find a collision-free new state using the masked action. Although calculating this maximum step size for a given action is always feasible, we need to compute the action mask before obtaining the final action and use it to influence the agent's planning. This implies the need to calculate the collision-free $v$ for all given angles.

\subsubsection{Efficient estimation of action mask}
\label{sec: action mask}
 We first introduce a vectorized obstacle distance representation $l_t$ at timestamp t, where the $i^{th}$ element $l_t[i]$ is the nearest obstacle distance at angle $\omega_i=i \cdot \Delta \omega$ in the ego coordinate and $\Delta \omega$ is the angular resolution. The movement distance during a certain period $\Delta t$ with some speed $v$ is denoted as step size $v\Delta t$.
 Consider the envelope area covered by the vehicle traveling with a steering angle $\delta_j$ and step size $v\Delta t$,
 denoted as $\text{S}_{\mathcal{E}}(\delta_j, v\Delta t)$. 
 Let $\mathcal{E}_{ij}(v)$ denote the distance from the envelope boundary to the origin in the ego vehicle's coordinate system at the $i^{th}$ angle $\omega_i$:
\begin{equation}
\begin{aligned}
    \mathcal{E}_{ij}(v) = max_e\{\|e\|_2  &\mid  e=(x,y) \in \text{S}_{\mathcal{E}}(\delta_j, v\Delta t),\\
    &x \operatorname{sin}(\omega_i) = y \operatorname{cos}(\omega_i) \}.
\end{aligned}
\end{equation}
 The collision constraint can then be expressed as $\mathcal{E}_{ij}(v) \le l_t[i]$. Then, the action mask calculation is equivalent to the following problem by introducing a new 2-dim variable $l$:
\begin{equation}
\begin{aligned}
    & \operatorname{min} \ l_{ij}; \ \operatorname{s.t.}\ l_{ij} = \mathcal{E}_{ij}(v),\ l_{ij} \le l_t[i].
\end{aligned}
\end{equation}
 While solving for the optimal $l_{ij}$ considering only obstacle point at $i^{th}$ angle $\omega_i$ and vehicle steering angle $\delta_j$ is equivalent to obtaining the maximum velocity, denoted as $v^*_{ij} = \mathcal{E}_{ij}^{-1}(l_t[i])$,
 calculating all envelope distances corresponding to each $i$ and $j$ in each interaction is computationally expensive, and the inverse of $\mathcal{E}_{ij}$ may not always exist. To deal with this issue, we propose the pre-calculation of the anchoring distance $\hat{l}_{ij}$ for $K_v$ discretized velocities:
\begin{equation}
\begin{aligned}
    & \hat{l}_{ij}[k] = \mathcal{E}_{ij}(\hat{v}[k]),\ \hat{v}[k] = v_{max} \frac{k}{K},\ k=0,1,...K_v.
\end{aligned}
\end{equation}
Then, we can obtain the upper and lower bounds of the masked velocity $v^*_{j}$ considering all obstacle points at steering angle $\delta_j$:
\begin{equation}
\label{eq: am calc}
\begin{aligned}
    & \hat{v}[k^*_{j}] \le v^*_{j} < \hat{v}[k^*_{j} +1], \\
     k^*_j = &\operatorname{min}_i \operatorname{max} \{k:\ l_t[i] - \hat{l}_{ij}[k] > 0 \}.
\end{aligned}
\end{equation}

We take $v^*_{j}=\hat{v}[k^*_{j}]$ as a conservative estimation of max step velocity, and all actions at discretized angle $\delta_j$ are calculated in a vectorized manner at once. Since the anchoring distance $\hat{l}_{ij}$ is independent of information of obstacles at timestamp $t$, a matrix of all anchoring distance $\hat{\mathcal{L}}$ can be pre-calculated before the training process starts. 
This implies that, as shown in Equation \ref{eq: am calc}, calculating the action mask at each interaction step is simplified to a comparison between two matrices.
 
\subsubsection{Combine action mask with agent's policy}
\label{sec: action mask 2}
The action mask process can be expressed as a function $f_{AM}$ from raw action to masked action:
\begin{equation}
    \hat{a}_t = f_{AM}(\pi_h(s_t)) = h(f_{AM}(\pi_\theta(s_t)), \pi_{RS}(s_t)).
    \label{eq: action mask 1}
\end{equation}
Here $f_{AM}$ is applied only to $\pi_\theta$ because $\pi_{RS}$ provides action only when a feasible RS curve exists. The action mask is also utilized to influence the probability distribution of actions. We here use $f_{am}: \mathcal{A} \rightarrow [0,1]$ to denote the maximum step size calculated by the action mask, where $f_{am}(a)=p$ indicates the maximum safe step size is $p\cdot v_{max} \triangle t$. Noticed that $f_{am}(a)$ can serve as a prior probability of $Collide(s_{t+1}) = False$ when $a \sim [a_{min}, a_{max}]$, the action mask can be applied on the distribution of actions from raw network output:
\begin{equation}
    log P(a_t|s_t) = \text{SoftMax}(log(\pi_\theta(a_t|s_t)) + log f_{am}(a_t)).
    \label{eq: action mask 2}
\end{equation}
 The $\text{SoftMax}$ here is the probability normalization operation, and $\pi_\theta(a_t|s_t)$ in practice we use a gaussian distribution:
\begin{equation}
    log(\pi_\theta(a|s)) = -\frac{(a-a_{mean})^2}{2 a_{std}^2} - log\left( \sqrt{2 \pi} a_{std} \right),
\end{equation}
where $a_{mean}$ is obtained by actor network using input $s$ and $a_{std}$ is a learnable parameter, as shown in Figure \ref{fig:overall_structure}. Equation \ref{eq: action mask 2} shows that the action mask can adjust the probability distribution of actions and avoid invalid actions by taking $f_{am}(a_t)=0$. We also utilize the post-process in Equation \ref{eq: action mask 1} to clip the velocity into a collision-free range in practice.

\section{Experiment}
\subsection{Scenario difficulty ranking}
\label{sec: ranking}
\begin{figure}[thb]
  \centering
  \includegraphics[width=\linewidth]{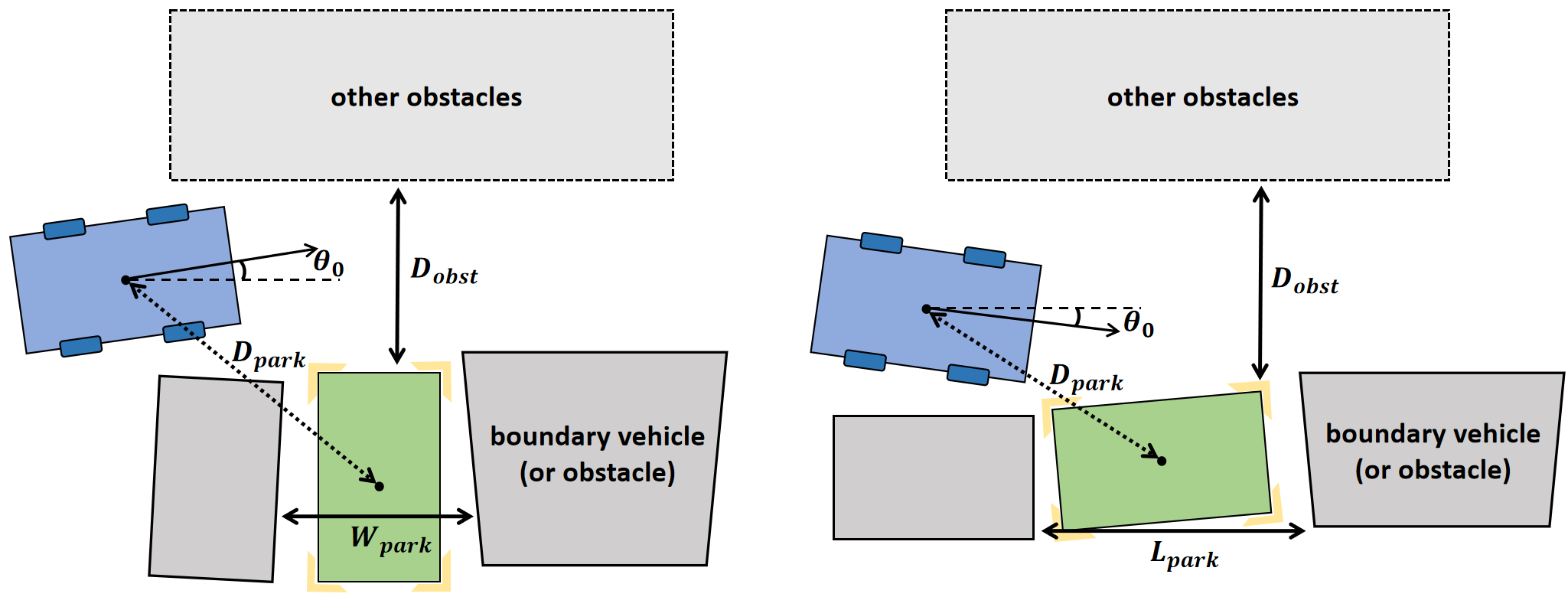}
  \caption{The description for ranking parameters in vertical parking (left) and parallel parking (right).}
  \label{fig: scenario describe}
\end{figure}

To better train and evaluate our approach, we rank the difficulty for parking scenarios into normal, complex, and extreme levels based on existing standards \textit{ISO 20900} and \textit{GB/T 41630-2022} \cite{ISO,GBT}. Generally, a parking space is defined by two boundary parking vehicles (or boundary obstacles) positioned on the two sides aligned along the curb-side edge of the road, as shown in Figure \ref{fig: scenario describe}. The boundaries define the parking space's length $L_{park}$ and width $W_{park}$.
In our setup, other obstacles are located at least $D_{obst}$ away from the parking spots to leave the drivable space.
We denote the parking vehicle's width as $W$, length as $L$, and the distance from the start to the target position as $D_{park}$. Table \ref{tab: difficulty} shows the categorization based on the aforementioned parameters, with narrower parking lots classified as more difficult.
Since parallel parking is more challenging than vertical parking in practice, we introduce an extreme difficulty level for it \cite{GBT}. 
Besides, since larger parking distances require more maneuvering over a longer distance and increase the number of obstacles encountered along the way, we rank the scenarios that $D_{park} > 15.0$ the complex scenario.
Note that we do not specify the initial orientation or position of the vehicle, meaning the starting conditions for the vehicle can be any collision-free configuration, which adds to the difficulty and diversity of scenarios. 

\begin{table}[bht]
   \caption{Parking scenario difficulty}
  \centering
  \label{tab: difficulty}
  \begin{tabular}{ l|l| c }
    \hline
    \textbf{Difficulty}& \multicolumn{1}{c|}{\textbf{Parallel}}&\textbf{Vertical} \\
    \hline
\multirow{2}{*}{Normal} &\textbf{P (N):} \quad $D_{obst} > 4.5 $  &\textbf{V (N):} \ $D_{obst}>7.0 $ \\
& $L_{park} > max(L+1.0, 1.25L)$ & $W_{park} > W+0.85$ \\

\hline
\multirow{2}{*}{Complex}&\textbf{P (C):} \quad $D_{obst} > 4.0$  &\textbf{V (C):} \ $D_{obst}>6.0 $ \\
& $L_{park} > max(L+0.9, 1.2L)$ & $W_{park} > W+0.4$ \\
\hline
\multirow{2}{*}{Extreme}&\textbf{P (E):} \quad  $D_{obst} > 3.5$ & \multirow{2}{*}{/} \\
& $L_{park} > max(L+0.6, 1.1L)$ &  \\
\hline
  \end{tabular}
  
  \rule{0pt}{6pt}
  V: vertical,
  P: parallel.
  N, C, and E in the bracket indicate the parking scenario difficulty, i.e., normal, complex, and extreme.
  unit: meter
\vspace{-8pt}
\end{table}

Based on the difficulty ranking method, scenarios can be generated and categorized using the pre-defined parameters. In our work, the scenarios are derived from two components:
\begin{itemize}
    \item Random generation using simulator: As shown in Figure \ref{fig: scenario describe}, obstacles and the parking spot in a scenario can be represented by multiple randomized parameters. Obstacles may include other stationary vehicles or irregular polygonal obstacles. We randomly set the initial orientation of the start with a Gaussian distribution with $mean(\theta_0) = 0$ and $std(\theta_0) = \pi/6$, and the initial position can be anywhere without collision between parking spots and other obstacles.
    \item Real-world scenario dataset: We utilize the Dragon Lake Parking (DLP) dataset to construct the DLP scenarios in our simulator \cite{shen2022parkpredict+}. This dataset is constructed from 3.5 hours of video data collected by a drone in a large parking lot, covering an area with about 400 parking spots and 5188 vehicles.
    While the original dataset was designed for intention and motion prediction tasks, we filtered out non-parking trajectories and dynamic interfering vehicles to obtain 253 static parking scenarios. The starting positions in these parking scenarios are randomly initialized along the recorded paths of the vehicles. These scenarios can be categorized into vertical parking with normal and complex difficulty levels.
\end{itemize}


\subsection{Implement details}
We conducted experiments in Tactics2D, an open-source simulator for driving decision-making \cite{li2023tactics2d}. 
This simulator provides sensor simulations, including lidar and bird's-eye view (BEV). 
In each episode, the simulator independently and randomly initializes the parallel or vertical parking scenarios of three difficulty levels generated by the simulator or from the DLP dataset.
We conducted 100,000 episodes of training in total and tested 2,000 trials in each scenario category, with no overlap between training and testing scenarios. Each time the scenario and its parameters are randomly generated according to the ranking method specified in section \ref{sec: ranking}
Details about hyperparameters for algorithm and simulation are listed in Table \ref{tab: param1} in the appendix.

\subsection{Results}

\subsubsection{Comaparion with baselines}
We compare the proposed method with the following baselines:
\begin{itemize}
    \item RS method \cite{reeds1990optimal}: The RS method is a classical geometric-based approach that calculates possible path types to obtain feasible routes. In our experiments, we enhance the original RS method by utilizing all calculated paths instead of only the shortest one.
    \item Hybrid A* \cite{dolgov2010path}: Hybrid A* is a search-based planning method incorporating heuristics considering obstacles and vehicle's non-holonomic constraints during node search. This method is widely applied in various planning tasks in the autonomous driving domain.
    \item Experience-based heuristic search method (EBHS) \cite{bernhard2018experience}: The EBHS leverages reinforcement learning to train a Q-network that serves as a heuristic function within the heuristic search algorithm. This learning-based approach allows the algorithm to derive more suitable heuristic functions from data and improve the exploring efficiency.
    \item Reinforcement learning baselines: When applying reinforcement learning methods to a specific task, a common practice is to establish a correspondence between the task elements and reinforcement learning components, using a deep network for function approximation. We explored this naive approach in our work, and experiments revealed that both PPO and SAC, in this manner, perform poorly in our parking path planning task. 
\end{itemize}

\begin{table}[thb]
   \caption{Planning success rate in different scenarios}
  \centering
  \label{tab: main compare}
  \begin{tabular}{ l|c c c c c c c}
    \hline
    \textbf{Algorithms}& \textbf{V(N)}&\textbf{P(N)}&\textbf{V(C)}&\textbf{P(C)}&\textbf{P(E)} &\textbf{D(N)} &\textbf{D(C)} \\
    \hline
RS & 36.9 & 10.4 & 30.4 & 1.5 & 0.3 & 4.0 & 0.1 \\
Hybrid A$*$  & 99.4 & 90.2 & 99.2 & 60.2 & 16.8 & 98.7 & 85.6 \\
EBHS & 96.4 & 95.3 & 92.4 & 89.1 & 43.2 & 91.8 & 61.4\\
PPO  & 93.2 & 74.2 & 82.9 & 69.0 & 58.4 & 65.2 & 34.2 \\
SAC  & 93.8 & 33.7 & 92.9 & 29.6 & 18.9 & 33.3 & 32.7  \\
HOPE(PPO)  & \textbf{100.0} & 99.4 & 99.8 & 97.5 & 94.2 & \textbf{99.5} & 97.6 \\
HOPE(SAC)  & \textbf{100.0} & \textbf{99.7} & \textbf{100.0} & \textbf{99.4} & \textbf{97.5} & 99.4 & \textbf{98.0}  \\
\hline
  \end{tabular}
  
  \rule{0pt}{6pt}
  D: DLP scenarios.
  unit: percentage
  \vspace{-8pt}
\end{table}

\begin{figure}[b]
  \centering
  \subfloat[]{\includegraphics[width=0.24\textwidth]{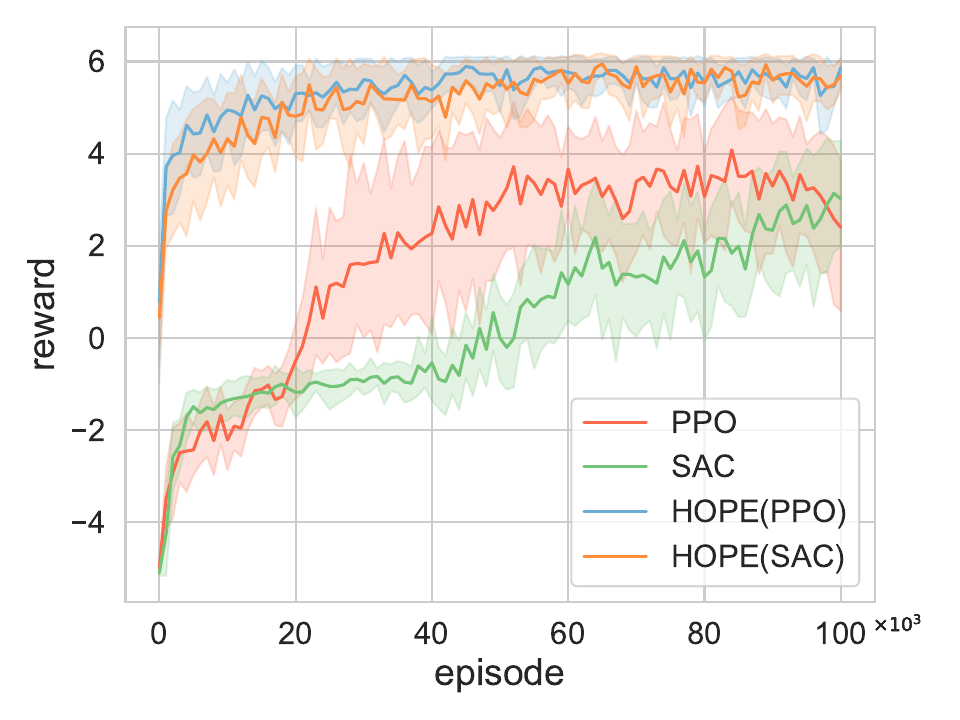}}\hfill
  \subfloat[]{\includegraphics[width=0.24\textwidth]{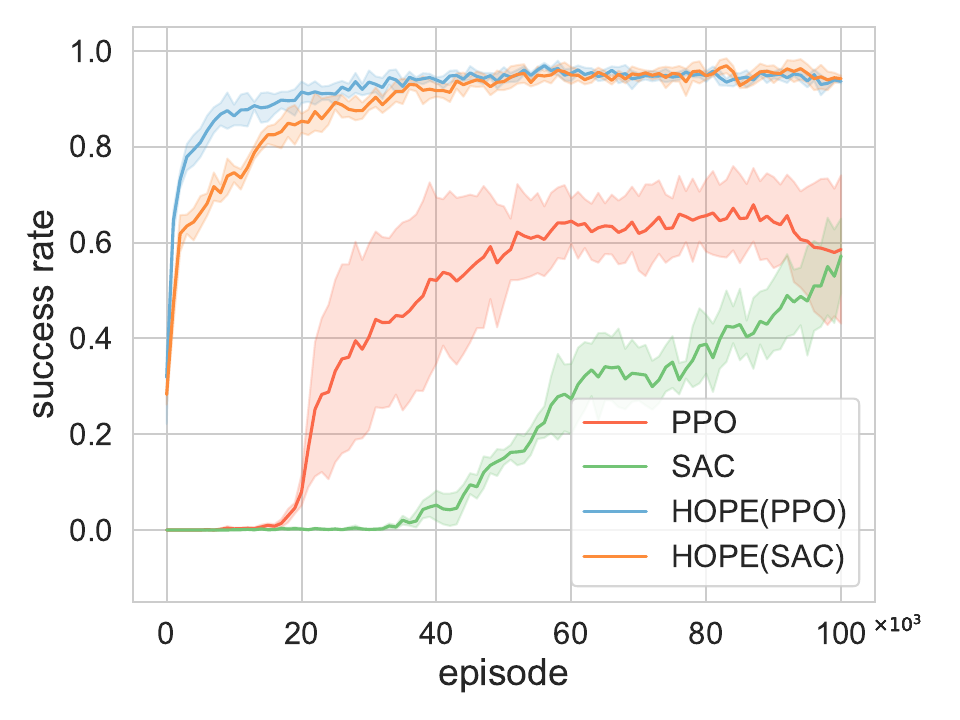}}\hfill
  \caption{The episode reward curves (a) and success rate (b)}
  \label{fig: training curve}
\end{figure}

Table \ref{tab: main compare} presents the planning success rates of these methods across different scene types and difficulty levels. We have separately listed the results for DLP scenarios since they originate from real-world parking environments. As shown in the table, a direct implementation of the RS method fails to provide a feasible path in most cases. Hybrid A* achieves a success rate of over 90\% in both normal scenarios and complex vertical parking cases. 
However, as the scene difficulty increases, its planning success rate significantly declines. In the extreme parallel parking scenarios, the most narrow cases, its success rate is only 16.8\%.
EBHS shows a significantly higher success rate in parallel parking scenarios than Hybrid A*. Specifically, in complex and extreme parallel parking scenarios, EBHS improves the success rate by nearly 30\% over Hybrid A*. This improvement is attributed to the learning-based Q-function in EBHS, which can better estimate the state values in these scenarios compared to the rule-based heuristic function in Hybrid A*. However, the overall success rate of EBHS is still not high enough to be considered robust.
The reinforcement learning baselines perform better than the rule-based RS method but fail to surpass the Hybrid A*. Their low success rates in several scenarios demonstrate that achieving good performance through overfitting in a few cases does not necessarily guarantee the trained agent's generalization capability. In contrast, our proposed HOPE, no matter based on on-policy PPO or off-policy SAC, outperforms all baselines and reaches a success rate of over 99.4\% in all normal scenarios and over 94\% in all scenarios.
Figure \ref{fig: training curve} shows the reward and success rate curves in the training process. The proposed method significantly improves training efficiency and success rate over the naive RL method by combining the RL agent and the RS policy.

\subsubsection{Further comparison with Hybrid A*}

\begin{figure*}[htb]
  \centering
  \captionsetup[subfigure]{labelformat=empty}
  \subfloat[(a-1)]{\includegraphics[width=0.19\textwidth]{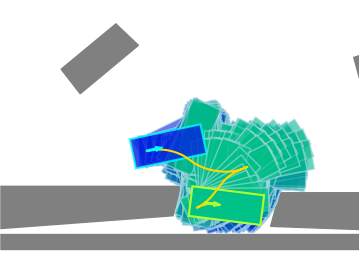}} \hfill
  \subfloat[(b-1)]{\includegraphics[width=0.19\textwidth]{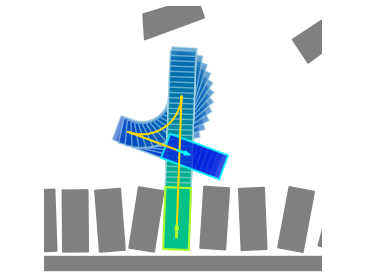}} \hfill
  \subfloat[(c-1)]{\includegraphics[width=0.17\textwidth]{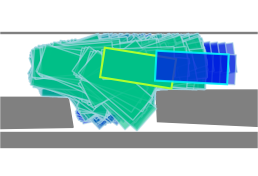}} \hfill
  \subfloat[(d-1)]{\includegraphics[width=0.18\textwidth]{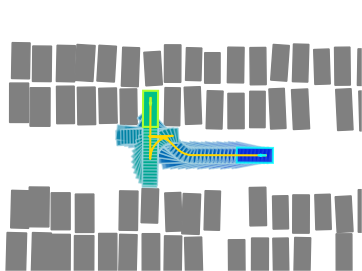}} \hfill
  \subfloat[(e-1)]{\includegraphics[width=0.19\textwidth]{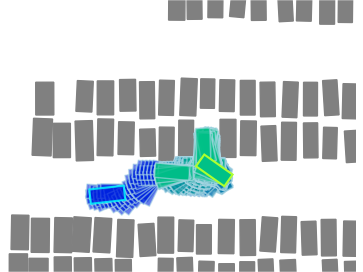}}
  \newline
  \subfloat[(a-2)]{\includegraphics[width=0.19\textwidth]{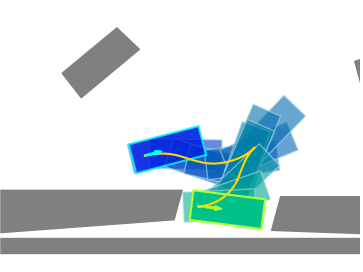}} \hfill
  \subfloat[(b-2)]{\includegraphics[width=0.19\textwidth]{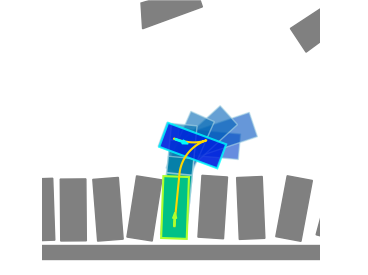}} \hfill
  \subfloat[(c-2)]{\includegraphics[width=0.17\textwidth]{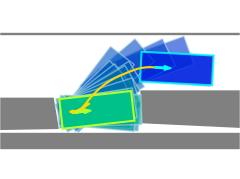}} \hfill
  \subfloat[(d-2)]{\includegraphics[width=0.18\textwidth]{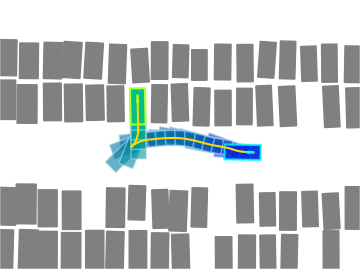}} \hfill
  \subfloat[(e-2)]{\includegraphics[width=0.19\textwidth]{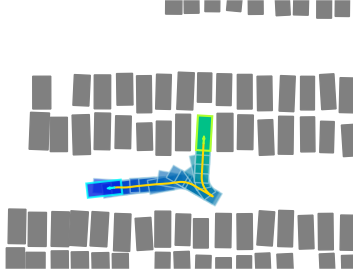}}
  \caption{The visualization of the planning process and results of the Hybrid A* (a-1)-(e-1) and the proposed HOPE (a-2)-(e-2). 
  The blue-to-green gradient rectangles represent the states explored during the algorithm's search process, while the yellow curves indicate the path planning result.
  In the normal parallel parking case shown in (a),  both methods provide concise path planning results. In the vertical parking scenario shown in (b) and the normal dlp scenario in (d), although both methods succeed in planning, our approach yields more reasonable results. In the narrow parallel parking scenario (c) and the scenario requiring parking with the front of the vehicle facing inward (e), the Hybrid A* fails to plan, while our approach succeeds.}
  \label{fig: case study}
\end{figure*}

As the Hybrid A* is a widely used method to this day, we further compare our approach with it through specific case studies. As shown in Figure \ref{fig: case study}, while both two methods can mask successful path planning in some cases, our method is capable of providing more concise and reasonable planning results, such as in Figure \ref{fig: case study} (b) and (d). 
As the parking space in the scene becomes narrow, Hybrid A* fails to explore a feasible path, as shown in Figure \ref{fig: case study} (c) and (e). In contrast, our method, through training, achieves the ability to maneuver within tight parking spaces and overcome local optima situations

\subsubsection{Computational consumption}
The average time consumption for a single-step prediction is 8.5 ms and can be broken down as follows: 2.7 ms for a single-step network forwarding, 2.8 ms for action mask calculation, and 3.0 ms for RS curve calculation. 
The simulator takes 8.3 ms each step for kinematics simulation and other information rendering.
The total time required to generate a complete planning result, including algorithmic computation and simulator simulation overhead, is shown in Table \ref{tab: time}.

\begin{table}[htb]
  \centering
   \caption{Time cost of complete path generation on average}
  \label{tab: time}
  \begin{tabularx}{\linewidth}{ c|X X X X X X X}
    \hline
    \textbf{Alg.}& \textbf{V(N)}&\textbf{P(N)}&\textbf{V(C)}&\textbf{P(C)}&\textbf{P(E)} &\textbf{D(N)} &\textbf{D(C)} \\
    \hline
    HOPE(PPO) & 314.6 & 451.6 & 369.1 & 549.8 & 891.4 & \textbf{433.0} & 699.0  \\
    HOPE(SAC) & \textbf{304.4} & \textbf{372.3} & \textbf{328.0} & \textbf{476.6} & \textbf{638.4} & 464.8 & \textbf{633.2}   \\
    \hline
  \end{tabularx}
  
  \rule{0pt}{6pt}
\textit{unit: microsecond(ms)}
\vspace{-15pt}
\end{table}

\subsection{Ablation studies}
\subsubsection{Hybrid policy with RS-curve}
We designed experiments to investigate how the RS method usageinfluences the performance of the hybrid policy. A hyperparameter in the hybrid strategy is the threshold distance \(d_{\text{rs}}\). The RS policy is considered only when the vehicle is closer to the target position than \(d_{\text{rs}}\). The results show that even reducing \(d_{\text{rs}}\) to 1 m has less than a 5\% impact on the success rate while increasing \(d_{\text{rs}}\) does not lead to a significant improvement. This result indicates that while RS curves assist in training the hybrid policy, the reinforcement learning agent does not overly rely on the RS method, allowing it to outperform rule-based approaches.
\begin{table}[bht]
  \centering
   \caption{Experiment on threshold distance for RS curve}
  \label{tab:exp6}
  \begin{tabularx}{\linewidth}{c|c|c X c X X X X}
    \hline
    \textbf{Alg.}&\textbf{$d_{rs}$}& \textbf{V(N)}&\textbf{P(N)}&\textbf{V(C)}&\textbf{P(C)}&\textbf{P(E)} &\textbf{D(N)} &\textbf{D(C)} \\
    \hline
\multirow{4}{*}{\parbox{0.7cm}{HOPE\\(PPO)}}& 1 & 99.2 & 97.4 & 95.4 & 95.0 & 92.6 &97.1 &95.0 \\
&5 & 99.6 & 98.7 & 97.6 & 96.1 & 93.2 & 99.5 & 96.7 \\
&10 & \textbf{100.0} & \textbf{99.4} & \textbf{99.8} & \textbf{97.5} & \textbf{94.2} & 99.5 & 97.6 \\
&15 & \textbf{100.0} & 99.3 & \textbf{99.8} & 96.6 & \textbf{94.2} & \textbf{99.9} & \textbf{98.4} \\
\hline
\multirow{4}{*}{\parbox{0.7cm}{HOPE\\(SAC)}}& 1 & 99.8 & 97.8 & 99.4 & 98.9 & 97.8 &95.8 &89.1 \\
&5 & \textbf{100.0} & 98.0 & 99.8 & 99.2 & 97.7 & 99.4 & 97.3 \\
&10 & \textbf{100.0} & \textbf{99.7} & \textbf{100.0} & \textbf{99.4} & 97.5 & \textbf{99.4} & 98.0 \\
&15 & \textbf{100.0} & 99.4 & \textbf{100.0} & 99.0 & \textbf{98.2} & 99.0 & \textbf{98.1} \\
\hline
  \end{tabularx}
  \vspace{-8pt}
\end{table}

\begin{figure}
    \centering
    \centering
  \subfloat[]{\includegraphics[width=0.24\textwidth]{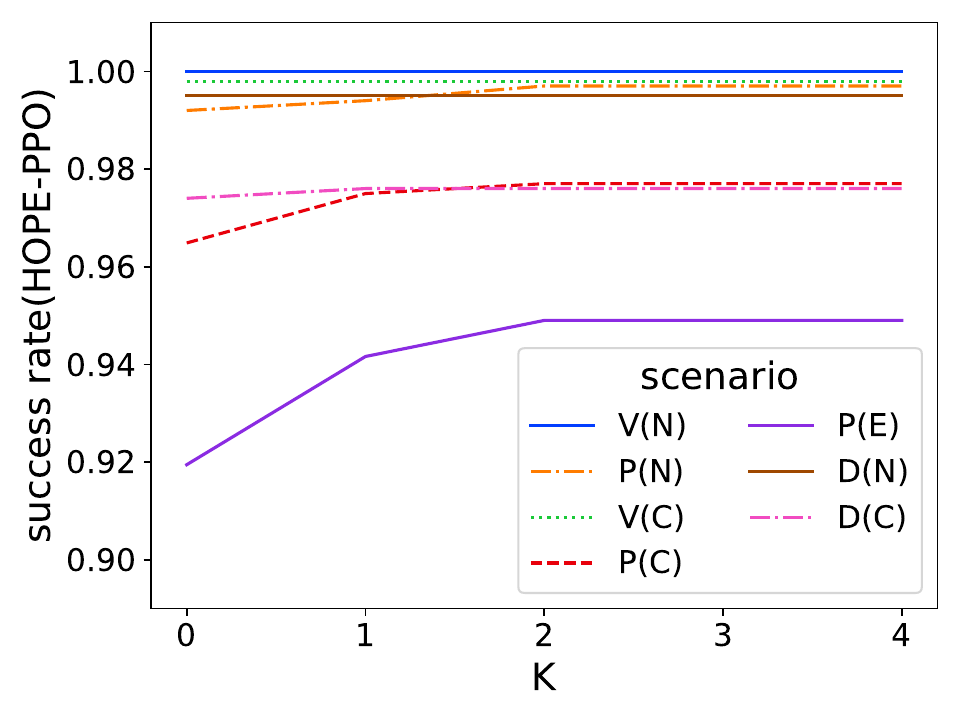}}\hfill
  \subfloat[]{\includegraphics[width=0.24\textwidth]{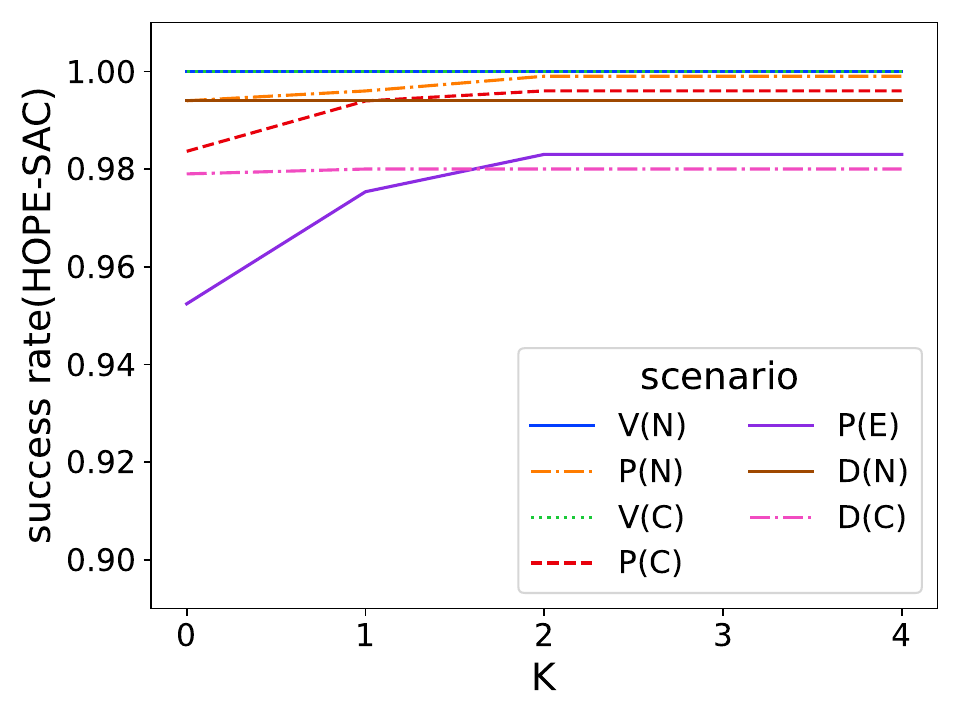}}\hfill
    \caption{Success rate using the shortest K RS curves.}
    \label{fig: exp k}
\end{figure}

Besides, while the original RS method only utilizes the shortest path, the shortest $K$ paths are considered in the hybrid policy. In practice, we choose $K=2$ to avoid redundant computations. At this point, the algorithm's performance has nearly saturated, as shown in Figure \ref{fig: exp k}.

\subsubsection{Other proposed modules}
More ablation experiment results are presented in Table \ref{tab: ablation}.
In the absence of the action mask mechanism, the success rate declines by more than 20\% in complex parallel parking scenarios and 30\% in extreme scenarios. This significant drop in performance indicates that utilizing an action mask to influence and constrain the agent's action is crucial for training effectiveness.
Besides, replacing the transformer with a multi-layer perceptron leads to a decline of more than 10\% in several scenarios, demonstrating the importance of multi-modal information fusion.
In situations where BEV data is unavailable, our model can still work effectively by removing the BEV-related input branch from the transformer module. This adjustment does not lead to significant performance drops, and in some scenarios, it even achieves comparable performance to the original.
Using the auto-encoder to pre-train the image encoder can enhance the training for the PPO-based agent but has a relatively minor impact on the SAC-based agent.

\begin{table}[htb]
\vspace{-8pt}
  \centering
   \caption{Ablation experiment}
  \label{tab: ablation}
  \begin{tabularx}{\linewidth}{ c|c X c X X X X}
    \hline
    \textbf{Experiments}& \textbf{V(N)}&\textbf{P(N)}&\textbf{V(C)}&\textbf{P(C)}&\textbf{P(E)} &\textbf{D(N)} &\textbf{D(C)} \\
    \hline
HOPE(PPO) & \textbf{100.0} & \textbf{99.4} & 99.8 & \textbf{97.5} & \textbf{94.2} & \textbf{99.5} & \textbf{97.6} \\
\hline
W/O AMP & 96.9 & 94.7 & 96.0 & 70.1 & 62.4 & 92.8 & 75.4 \\
W/O TF & 99.8 & 98.5 & 99.7 & 93.3 & 82.8 & 99.4 & 92.1 \\
W/O AMI & 99.6 & 99.0 & 98.1 & 97.3 & 91.0 & 98.8 & 94.4 \\
W/O BEV & \textbf{100.0} & 98.8 & \textbf{99.9} & 96.1 & 92.8 & 99.4 & 93.5 \\
W/O AE & 99.9 & 98.2 & 99.5 & 95.4 & 86.2 & 97.0 & 85.7 \\
\hline
\hline
HOPE(SAC)  & \textbf{100.0} & 99.7 & \textbf{100.0} & \textbf{99.4} & 97.5 & 99.4 & 98.0  \\
\hline
W/O AMP & 96.5 & 49.1 & 96.3 & 36.5 & 23.4 & 78.2 & 57.2 \\
W/O TF & 99.8 & 90.6 & 98.8 & 82.3 & 72.1 & 97.8 & 87.9 \\
W/O AMI & 99.9 & 96.4 & 98.7 & 91.9 & 84.9 & 97.2 & 92.3 \\
W/O BEV & 99.8 & 97.8 & 98.7 & 97.9 & \textbf{97.6} & 98.1 & 96.5 \\
W/O AE & \textbf{100.0} & \textbf{99.8} & \textbf{100.0} & 99.0 & 96.7 & \textbf{99.7} & \textbf{99.0} \\
\hline
  \end{tabularx}
  
  \rule{0pt}{6pt}
W/O AMP: without the action mask in post-processing.  W/O TF: use MLP instead of transformer as the backbone.  W/O AMI: without the action mask in network input.  W/O BEV: without the BEV image in network input.  W/O AE: not employ auto-encoder to pre-train the image encoder.
\vspace{-8pt}
\end{table}

\subsubsection{Scenario difficulty}
We also examined the impact of restricting the training scenario difficulty on the agent's performance. As shown in Table \ref{tab: train difficulty}, training solely on normal cases leads to a 20\% to 30\% decline in success rate in extreme scenarios, even with the help of the RS policy and action mask mechanism. Generalization capability under all difficulties can only be achieved when the training scenarios are sufficiently complex and diverse, which also underscores the importance of ranking difficulty in scenario taxonomy.

  

\begin{table}[htb]
  \centering
   \caption{Experiments on training scenario difficulty}
  \label{tab: train difficulty}
  \begin{tabularx}{\linewidth}{ c  c | c c c c c c c}
    \hline
    \multicolumn{2}{c|}{\textbf{difficulty}}& \textbf{V(N)}&\textbf{P(N)}&\textbf{V(C)}&\textbf{P(C)}&\textbf{P(E)} &\textbf{D(N)} &\textbf{D(C)} \\
    \hline
\multirow{3}{*}{\parbox{0.7cm}{HOPE\\(PPO)}} & N & \textbf{100.0} & 98.4 & 99.8 & 89.5 & 64.6 & 99.8 & 67.5 \\
& +C & 99.9 & 98.5 & \textbf{99.9} & 94.8 & 76.1 & \textbf{99.9} & \textbf{98.1} \\
& +E & \textbf{100.0} & \textbf{99.4} & 99.8 & \textbf{97.5} & \textbf{94.2} & 99.5 & 97.6 \\
\hline
\multirow{3}{*}{\parbox{0.7cm}{HOPE\\(SAC)}} & N & \textbf{100.0} & 97.6 & \textbf{100.0} & 91.1 & 77.6 & 98.4 & 60.3 \\
& +C & \textbf{100.0} & 98.3 & \textbf{100.0} & 91.5 & 80.0 & \textbf{99.7} & 95.4 \\
& +E & \textbf{100.0} & \textbf{99.7} & \textbf{100.0} & \textbf{99.4} & \textbf{97.5} & 99.4 & \textbf{98.0} \\
\hline
  \end{tabularx}
  
  \rule{0pt}{6pt}
N: use only normal cases in training. +C: use normal and complex cases. +E: use cases of all difficulty levels.
\vspace{-15pt}
\end{table}

\subsection{Real-World Experiments}
We also conducted real-world experiments in an indoor parking garage to show the applicability of our method in the real world. Our experimental platform was a Changan CS55E-Rock vehicle with a dimension of $4.6m \times 1.9m$ and we fine-tuned our model with the vehicle's parameter in the same simulation environment. More details about the platform and the overall system can be found in the appendix \ref{appendix: real-world}.

\begin{figure}[htb]
  \centering
  \captionsetup[subfigure]{labelformat=empty}
  \subfloat[(a-1)]{\includegraphics[width=0.45\linewidth]{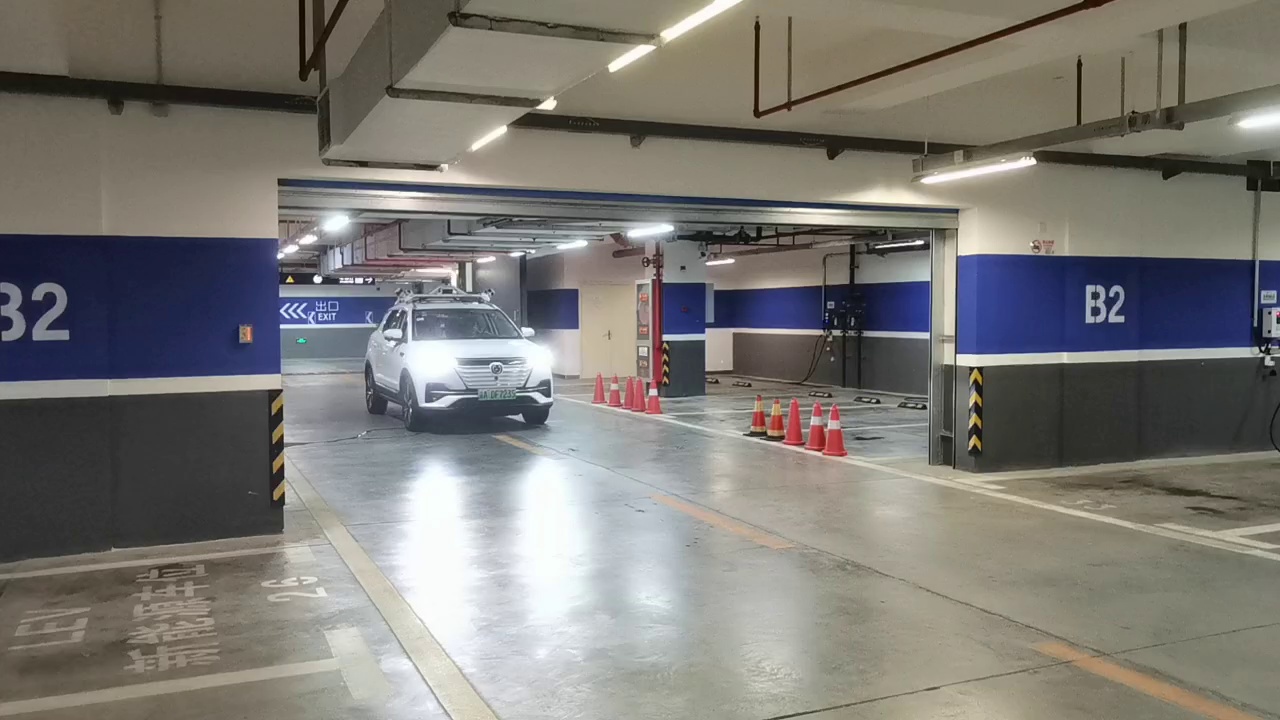}}
  \subfloat[(a-2)]{\includegraphics[width=0.45\linewidth]{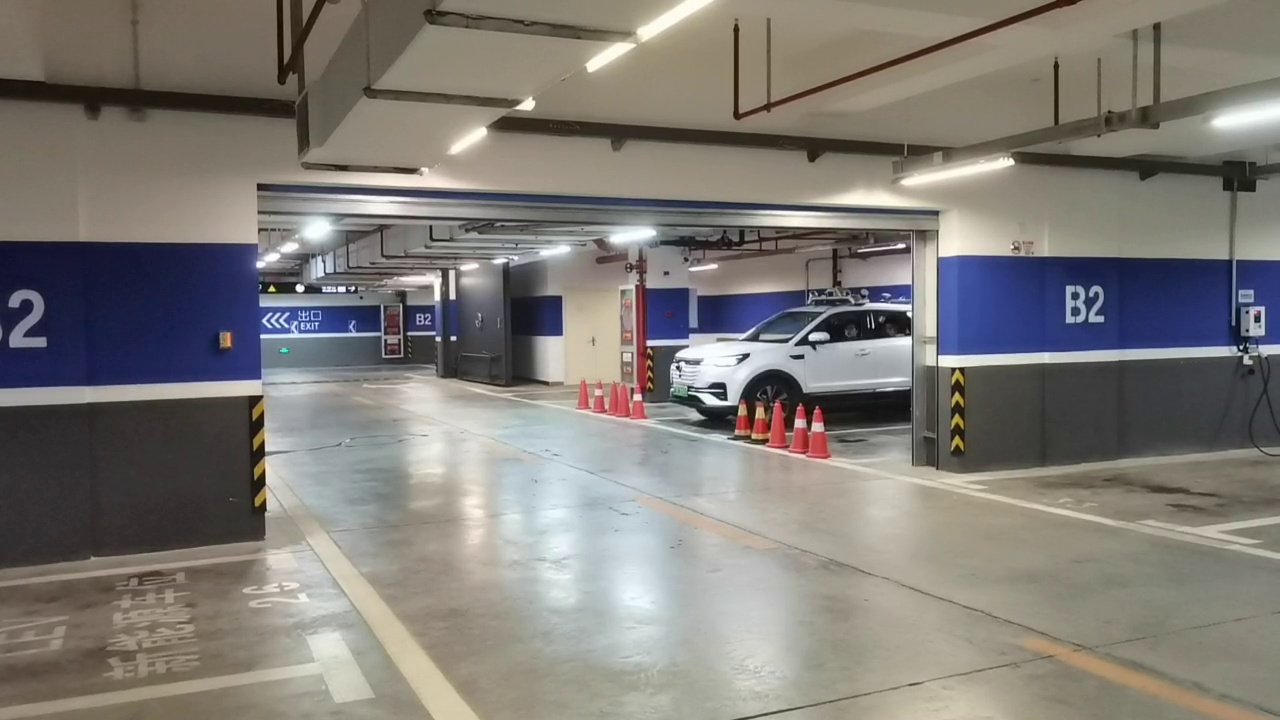}}
  
  \subfloat[(b-1)]{\includegraphics[width=0.45\linewidth]{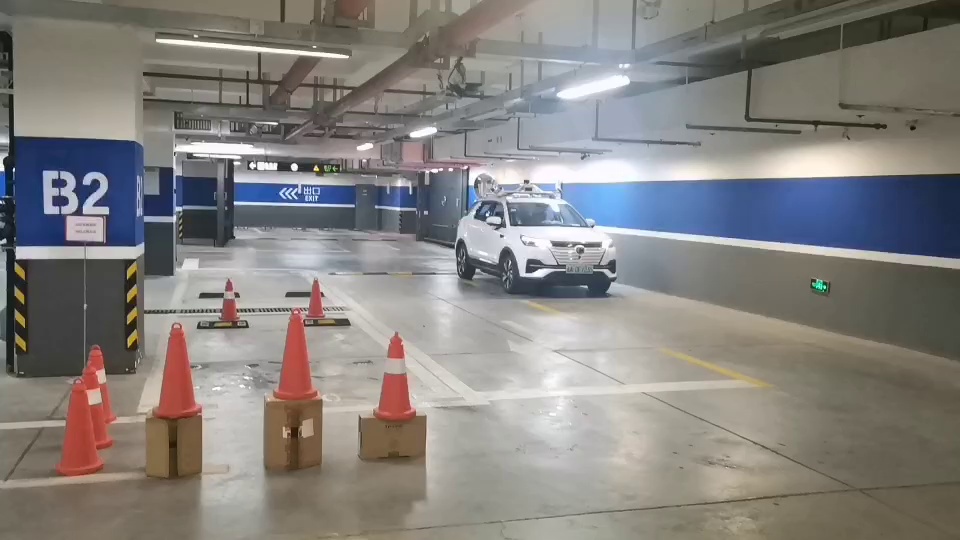}}
  \subfloat[(b-2)]{\includegraphics[width=0.45\linewidth]{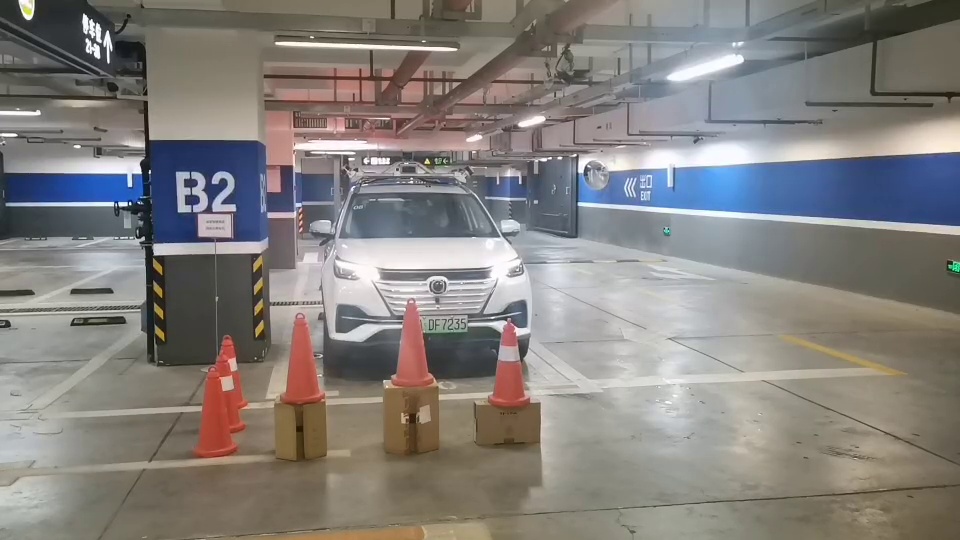}}
  
  \subfloat[(c-1)]{\includegraphics[width=0.45\linewidth]{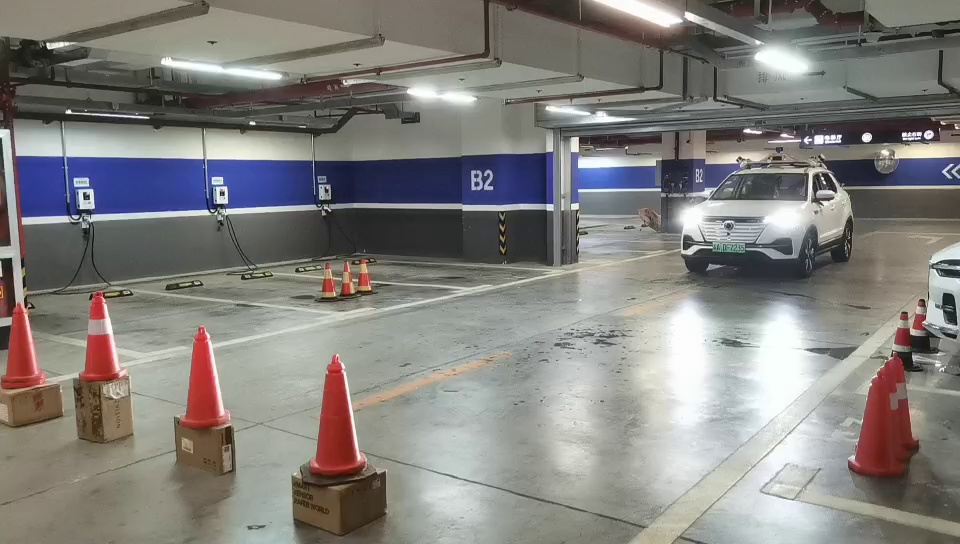}}
  \subfloat[(c-2)]{\includegraphics[width=0.45\linewidth]{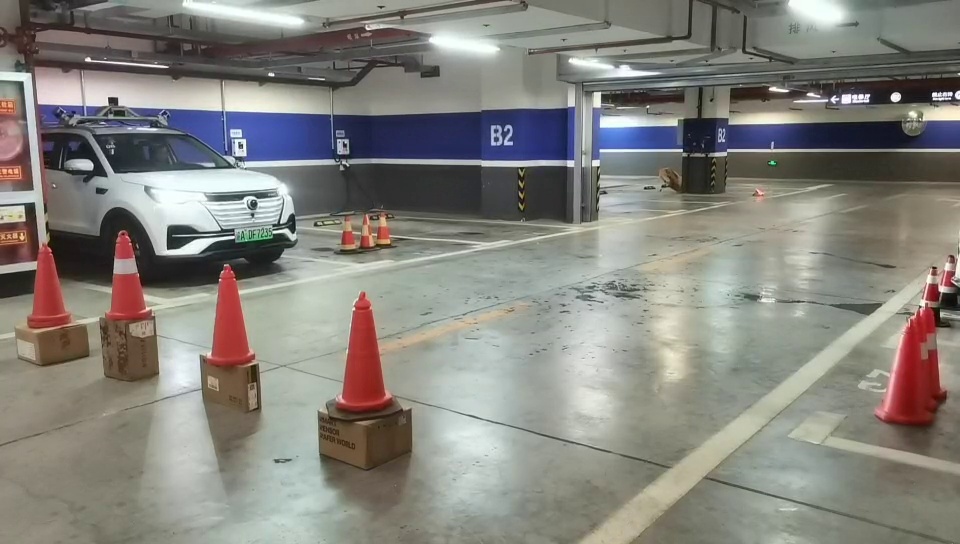}}
  \caption{The illustration of our real-world experiments in scenarios of vertical parking (a-1,2), parallel parking (b-1,2), and dead-end parking (c-1,2). The initial position is shown in the left images and the right images show the target parking lots where the vehicle is positioned. }
  \label{fig: realworld}
\end{figure}

We tested our method on vertical parking, parallel parking, and an additional challenging dead-end vertical parking scenario, as shown in Figure \ref{fig: realworld}. To test the generalization ability of our algorithm, these scenarios are not included in the training set. In all three scenarios, our algorithm successfully planned collision-free paths and guided the vehicle to complete the parking process. Detailed videos of the parking processes can be found at \url{https://www.youtube.com/watch?v=62w9qhjIuRI}. It is worth noting that the dead-end scenario in Figure \ref{fig: realworld} (c) represents a situation where the drivable area in front of the parking spot is narrow, with one side obstructed by obstacles. This requires the vehicle to perform extensive maneuvering and multiple gear shifts to adjust its orientation, making it a highly challenging scenario. These results demonstrate the practical applicability and generalization capability of our algorithm in real-world scenarios.

\section{Conclusion and Future Work}
In this paper, we proposed a reinforcement learning-based path planning method in diverse parking scenarios with static obstacles. Instead of directly training the parking agent with existing RL methods, we introduced a hybrid policy that integrates the RS curve and PPO as well as SAC. To enhance the training efficiency and performance of reinforcement learning agents, we introduce a method to calculate and implement the action mask mechanism in the task of parking path planning. A transformer-based network structure is utilized to fuse information about obstacles and target parking spots from different inputs and output the planning actions. 

To better train and evaluate the proposed method, we introduce a difficulty ranking approach for parking scenarios, which categorizes scenarios into normal, complex, and extreme based on obstacles and parking space conditions. We utilized a simulator and a real-world dataset to construct training and testing scenarios. Comprehensive experiments show the effectiveness of our method against other rule-based and RL-based methods. The results indicate that our approach outperforms the Hybrid A* in planning success rates across all types of scenarios, particularly in complex and extreme situations. 
Compared to the typical approach of directly using reinforcement learning, our proposed method combines the advantages of both rule-based and learning-based methods, resulting in an effective planner with generalization capabilities across different scenarios.


This paper shows that the learning-based approach can serve as a promising tool in dealing with complex and diverse parking scenarios. The proposed HOPE serves as a potential planner to deal with the intricate scenarios where the rule-based methods fail to plan a feasible path. 
Besides, since only static scenarios are considered in our work, our method can be integrated with a dynamic obstacle avoidance system to handle real-world participants to ensure compatibility and safety in dynamic environments. In the future, more efforts could be made to apply the learning-based method directly in parking scenarios with dynamic obstacles. To support this, enhancing simulation environments with interactive traffic participant models that realistically react to the autonomous vehicle would greatly improve the training and evaluation of such methods.

\bibliography{ref}

\begin{thebibliography}{10}
\providecommand{\url}[1]{#1}
\csname url@samestyle\endcsname
\providecommand{\newblock}{\relax}
\providecommand{\bibinfo}[2]{#2}
\providecommand{\BIBentrySTDinterwordspacing}{\spaceskip=0pt\relax}
\providecommand{\BIBentryALTinterwordstretchfactor}{4}
\providecommand{\BIBentryALTinterwordspacing}{\spaceskip=\fontdimen2\font plus
\BIBentryALTinterwordstretchfactor\fontdimen3\font minus \fontdimen4\font\relax}
\providecommand{\BIBforeignlanguage}[2]{{%
\expandafter\ifx\csname l@#1\endcsname\relax
\typeout{** WARNING: IEEEtran.bst: No hyphenation pattern has been}%
\typeout{** loaded for the language `#1'. Using the pattern for}%
\typeout{** the default language instead.}%
\else
\language=\csname l@#1\endcsname
\fi
#2}}
\providecommand{\BIBdecl}{\relax}
\BIBdecl

\bibitem{song2016analysis}
Y.~Song and C.~Liao, ``Analysis and review of state-of-the-art automatic parking assist system,'' in \emph{2016 IEEE International Conference on Vehicular Electronics and Safety (ICVES)}.\hskip 1em plus 0.5em minus 0.4em\relax IEEE, 2016, pp. 1--6.

\bibitem{wang2014automatic}
W.~Wang, Y.~Song, J.~Zhang, and H.~Deng, ``Automatic parking of vehicles: A review of literatures,'' \emph{International Journal of Automotive Technology}, vol.~15, pp. 967--978, 2014.

\bibitem{li2021optimization}
B.~Li, T.~Acarman, Y.~Zhang, Y.~Ouyang, C.~Yaman, Q.~Kong, X.~Zhong, and X.~Peng, ``Optimization-based trajectory planning for autonomous parking with irregularly placed obstacles: A lightweight iterative framework,'' \emph{IEEE Transactions on Intelligent Transportation Systems}, vol.~23, no.~8, pp. 11\,970--11\,981, 2021.

\bibitem{likmeta2020combining}
A.~Likmeta, A.~M. Metelli, A.~Tirinzoni, R.~Giol, M.~Restelli, and D.~Romano, ``Combining reinforcement learning with rule-based controllers for transparent and general decision-making in autonomous driving,'' \emph{Robotics and Autonomous Systems}, vol. 131, p. 103568, 2020.

\bibitem{teng2023motion}
S.~Teng, X.~Hu, P.~Deng, B.~Li, Y.~Li, Y.~Ai, D.~Yang, L.~Li, Z.~Xuanyuan, F.~Zhu \emph{et~al.}, ``Motion planning for autonomous driving: The state of the art and future perspectives,'' \emph{IEEE Transactions on Intelligent Vehicles}, 2023.

\bibitem{kiran2021deep}
B.~R. Kiran, I.~Sobh, V.~Talpaert, P.~Mannion, A.~A. Al~Sallab, S.~Yogamani, and P.~P{\'e}rez, ``Deep reinforcement learning for autonomous driving: A survey,'' \emph{IEEE Transactions on Intelligent Transportation Systems}, vol.~23, no.~6, pp. 4909--4926, 2021.

\bibitem{grigorescu2020survey}
S.~Grigorescu, B.~Trasnea, T.~Cocias, and G.~Macesanu, ``A survey of deep learning techniques for autonomous driving,'' \emph{Journal of Field Robotics}, vol.~37, no.~3, pp. 362--386, 2020.

\bibitem{reeds1990optimal}
J.~Reeds and L.~Shepp, ``Optimal paths for a car that goes both forwards and backwards,'' \emph{Pacific journal of mathematics}, vol. 145, no.~2, pp. 367--393, 1990.

\bibitem{vaswani2017attention}
A.~Vaswani, N.~Shazeer, N.~Parmar, J.~Uszkoreit, L.~Jones, A.~N. Gomez, {\L}.~Kaiser, and I.~Polosukhin, ``Attention is all you need,'' \emph{Advances in neural information processing systems}, vol.~30, 2017.

\bibitem{liang2012automatic}
Z.~Liang, G.~Zheng, and J.~Li, ``Automatic parking path optimization based on bezier curve fitting,'' in \emph{2012 IEEE International Conference on Automation and Logistics}.\hskip 1em plus 0.5em minus 0.4em\relax IEEE, 2012, pp. 583--587.

\bibitem{vorobieva2014automatic2}
H.~Vorobieva, S.~Glaser, N.~Minoiu-Enache, and S.~Mammar, ``Automatic parallel parking with geometric continuous-curvature path planning,'' in \emph{2014 IEEE Intelligent Vehicles Symposium Proceedings}.\hskip 1em plus 0.5em minus 0.4em\relax IEEE, 2014, pp. 465--471.

\bibitem{dubins1957curves}
L.~E. Dubins, ``On curves of minimal length with a constraint on average curvature, and with prescribed initial and terminal positions and tangents,'' \emph{American Journal of mathematics}, vol.~79, no.~3, pp. 497--516, 1957.

\bibitem{du2014autonomous}
X.~Du and K.~K. Tan, ``Autonomous reverse parking system based on robust path generation and improved sliding mode control,'' \emph{IEEE Transactions on Intelligent Transportation Systems}, vol.~16, no.~3, pp. 1225--1237, 2014.

\bibitem{kim2014auto}
J.~M. Kim, K.~I. Lim, and J.~H. Kim, ``Auto parking path planning system using modified reeds-shepp curve algorithm,'' in \emph{2014 11th International Conference on Ubiquitous Robots and Ambient Intelligence (URAI)}.\hskip 1em plus 0.5em minus 0.4em\relax IEEE, 2014, pp. 311--315.

\bibitem{dolgov2010path}
D.~Dolgov, S.~Thrun, M.~Montemerlo, and J.~Diebel, ``Path planning for autonomous vehicles in unknown semi-structured environments,'' \emph{The international journal of robotics research}, vol.~29, no.~5, pp. 485--501, 2010.

\bibitem{sedighi2019guided}
S.~Sedighi, D.-V. Nguyen, and K.-D. Kuhnert, ``Guided hybrid a-star path planning algorithm for valet parking applications,'' in \emph{2019 5th international conference on control, automation and robotics (ICCAR)}.\hskip 1em plus 0.5em minus 0.4em\relax IEEE, 2019, pp. 570--575.

\bibitem{sheng2021autonomous}
W.~Sheng, B.~Li, and X.~Zhong, ``Autonomous parking trajectory planning with tiny passages: a combination of multistage hybrid a-star algorithm and numerical optimal control,'' \emph{IEEE Access}, vol.~9, pp. 102\,801--102\,810, 2021.

\bibitem{czubenko2015autonomous}
M.~Czubenko, Z.~Kowalczuk, and A.~Ordys, ``Autonomous driver based on an intelligent system of decision-making,'' \emph{Cognitive computation}, vol.~7, pp. 569--581, 2015.

\bibitem{zhang2020optimization}
X.~Zhang, A.~Liniger, and F.~Borrelli, ``Optimization-based collision avoidance,'' \emph{IEEE Transactions on Control Systems Technology}, vol.~29, no.~3, pp. 972--983, 2020.

\bibitem{he2021tdr}
R.~He, J.~Zhou, S.~Jiang, Y.~Wang, J.~Tao, S.~Song, J.~Hu, J.~Miao, and Q.~Luo, ``Tdr-obca: A reliable planner for autonomous driving in free-space environment,'' in \emph{2021 American Control Conference (ACC)}.\hskip 1em plus 0.5em minus 0.4em\relax IEEE, 2021, pp. 2927--2934.

\bibitem{han2023efficient}
Z.~Han, Y.~Wu, T.~Li, L.~Zhang, L.~Pei, L.~Xu, C.~Li, C.~Ma, C.~Xu, S.~Shen \emph{et~al.}, ``An efficient spatial-temporal trajectory planner for autonomous vehicles in unstructured environments,'' \emph{IEEE Transactions on Intelligent Transportation Systems}, 2023.

\bibitem{liu2017parking}
W.~Liu, Z.~Li, L.~Li, and F.-Y. Wang, ``Parking like a human: A direct trajectory planning solution,'' \emph{IEEE Transactions on Intelligent Transportation Systems}, vol.~18, no.~12, pp. 3388--3397, 2017.

\bibitem{rathour2018vision}
S.~Rathour, V.~John, M.~Nithilan, and S.~Mita, ``Vision and dead reckoning-based end-to-end parking for autonomous vehicles,'' in \emph{2018 IEEE Intelligent Vehicles Symposium (IV)}.\hskip 1em plus 0.5em minus 0.4em\relax IEEE, 2018, pp. 2182--2187.

\bibitem{chai2020design}
R.~Chai, A.~Tsourdos, A.~Savvaris, S.~Chai, Y.~Xia, and C.~P. Chen, ``Design and implementation of deep neural network-based control for automatic parking maneuver process,'' \emph{IEEE Transactions on Neural Networks and Learning Systems}, vol.~33, no.~4, pp. 1400--1413, 2020.

\bibitem{chai2022deep}
R.~Chai, D.~Liu, T.~Liu, A.~Tsourdos, Y.~Xia, and S.~Chai, ``Deep learning-based trajectory planning and control for autonomous ground vehicle parking maneuver,'' \emph{IEEE Transactions on Automation Science and Engineering}, 2022.

\bibitem{ho2016generative}
J.~Ho and S.~Ermon, ``Generative adversarial imitation learning,'' \emph{Advances in neural information processing systems}, vol.~29, 2016.

\bibitem{song2022time}
S.~Song, H.~Chen, H.~Sun, M.~Liu, and T.~Xia, ``Time-optimized online planning for parallel parking with nonlinear optimization and improved monte carlo tree search,'' \emph{IEEE Robotics and Automation Letters}, vol.~7, no.~2, pp. 2226--2233, 2022.

\bibitem{mnih2015human}
V.~Mnih, K.~Kavukcuoglu, D.~Silver, A.~A. Rusu, J.~Veness, M.~G. Bellemare, A.~Graves, M.~Riedmiller, A.~K. Fidjeland, G.~Ostrovski \emph{et~al.}, ``Human-level control through deep reinforcement learning,'' \emph{nature}, vol. 518, no. 7540, pp. 529--533, 2015.

\bibitem{bernhard2018experience}
J.~Bernhard, R.~Gieselmann, K.~Esterle, and A.~Knoll, ``Experience-based heuristic search: Robust motion planning with deep q-learning,'' in \emph{2018 21st International conference on intelligent transportation systems (ITSC)}.\hskip 1em plus 0.5em minus 0.4em\relax IEEE, 2018, pp. 3175--3182.

\bibitem{du2020trajectory}
Z.~Du, Q.~Miao, and C.~Zong, ``Trajectory planning for automated parking systems using deep reinforcement learning,'' \emph{International Journal of Automotive Technology}, vol.~21, pp. 881--887, 2020.

\bibitem{yuan2023hierarchical}
Z.~Yuan, Z.~Wang, X.~Li, L.~Li, and L.~Zhang, ``Hierarchical trajectory planning for narrow-space automated parking with deep reinforcement learning: A federated learning scheme,'' \emph{Sensors}, vol.~23, no.~8, p. 4087, 2023.

\bibitem{althoff2017commonroad}
M.~Althoff, M.~Koschi, and S.~Manzinger, ``Commonroad: Composable benchmarks for motion planning on roads,'' in \emph{2017 IEEE Intelligent Vehicles Symposium (IV)}.\hskip 1em plus 0.5em minus 0.4em\relax IEEE, 2017, pp. 719--726.

\bibitem{schulman2017proximal}
J.~Schulman, F.~Wolski, P.~Dhariwal, A.~Radford, and O.~Klimov, ``Proximal policy optimization algorithms,'' \emph{arXiv preprint arXiv:1707.06347}, 2017.

\bibitem{haarnoja2018soft}
T.~Haarnoja, A.~Zhou, P.~Abbeel, and S.~Levine, ``Soft actor-critic: Off-policy maximum entropy deep reinforcement learning with a stochastic actor,'' in \emph{International conference on machine learning}.\hskip 1em plus 0.5em minus 0.4em\relax PMLR, 2018, pp. 1861--1870.

\bibitem{shao2023safety}
H.~Shao, L.~Wang, R.~Chen, H.~Li, and Y.~Liu, ``Safety-enhanced autonomous driving using interpretable sensor fusion transformer,'' in \emph{Conference on Robot Learning}.\hskip 1em plus 0.5em minus 0.4em\relax PMLR, 2023, pp. 726--737.

\bibitem{huang2020closer}
S.~Huang and S.~Onta{\~n}{\'o}n, ``A closer look at invalid action masking in policy gradient algorithms,'' \emph{arXiv preprint arXiv:2006.14171}, 2020.

\bibitem{ISO}
``Intelligent transport systems - partially-automated parking systems (paps) - performance requirements and test procedures,'' ISO 20900:2023, 2023.

\bibitem{GBT}
``Performance requirements and test methods for intelligent parking assist system,'' GB/T 41630-2022, 2022.

\bibitem{shen2022parkpredict+}
X.~Shen, M.~Lacayo, N.~Guggilla, and F.~Borrelli, ``Parkpredict+: Multimodal intent and motion prediction for vehicles in parking lots with cnn and transformer,'' in \emph{2022 IEEE 25th International Conference on Intelligent Transportation Systems (ITSC)}.\hskip 1em plus 0.5em minus 0.4em\relax IEEE, 2022, pp. 3999--4004.

\bibitem{li2023tactics2d}
Y.~Li, S.~Zhang, M.~Jiang, X.~Chen, and M.~Yang, ``Tactics2d: A multi-agent reinforcement learning environment for driving decision-making,'' \emph{arXiv preprint arXiv:2311.11058}, 2023.

\end{thebibliography}
\bibliographystyle{IEEEtran}

\begin{IEEEbiography}
[{\includegraphics[width=1in,height=1.25in,clip,keepaspectratio]{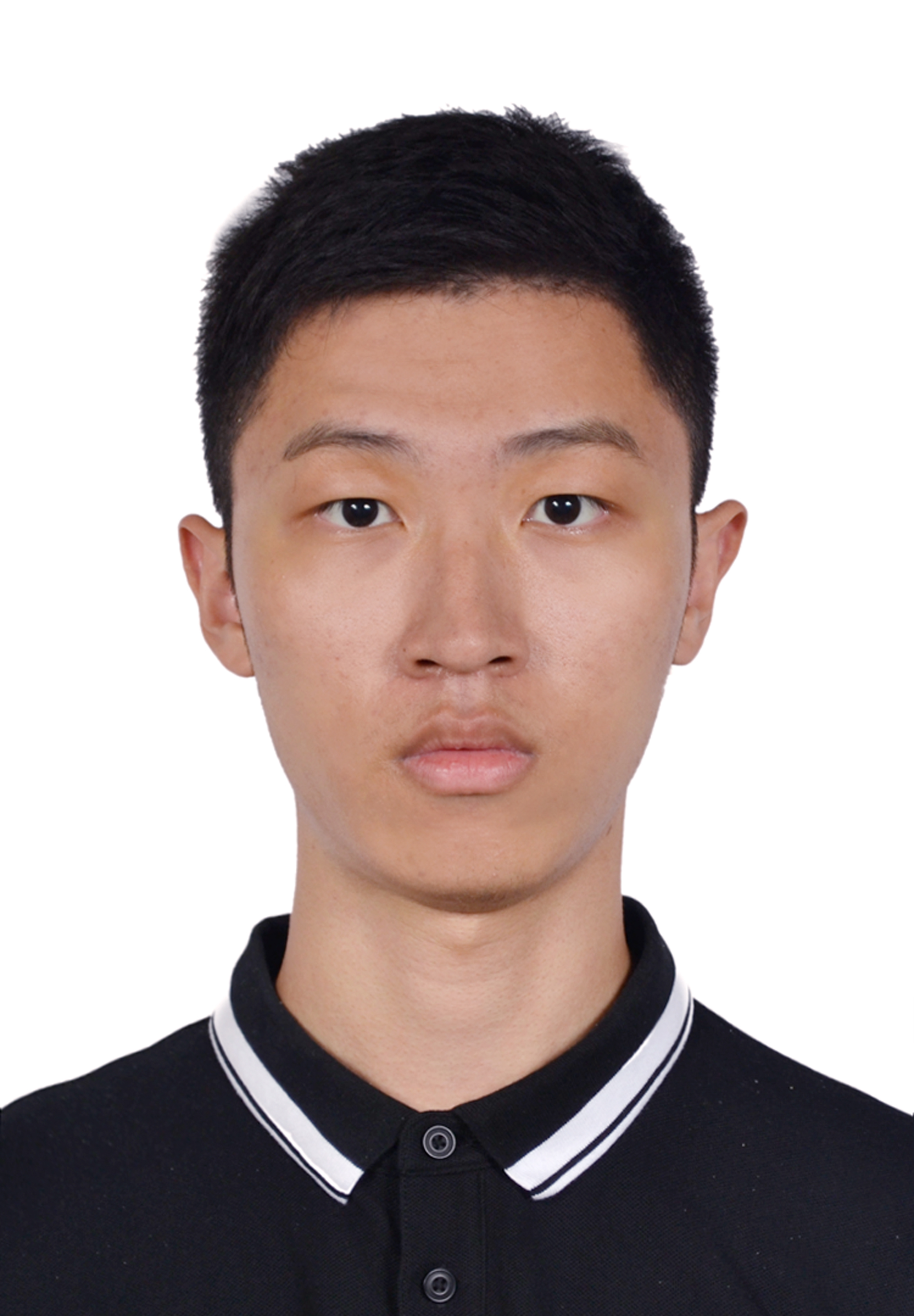}}]{Mingyang Jiang} 
    received a Bachelor's degree in engineering from Shanghai Jiao Tong University, Shanghai, China, in 2023. He is working towards a Master's degree in Control Science and Engineering from Shanghai Jiao Tong University. His main research interests are end-to-end planning, driving decision-making, and reinforcement learning for autonomous vehicles.
\end{IEEEbiography}

\begin{IEEEbiography}
[{\includegraphics[width=1in,height=1.25in,clip,keepaspectratio]{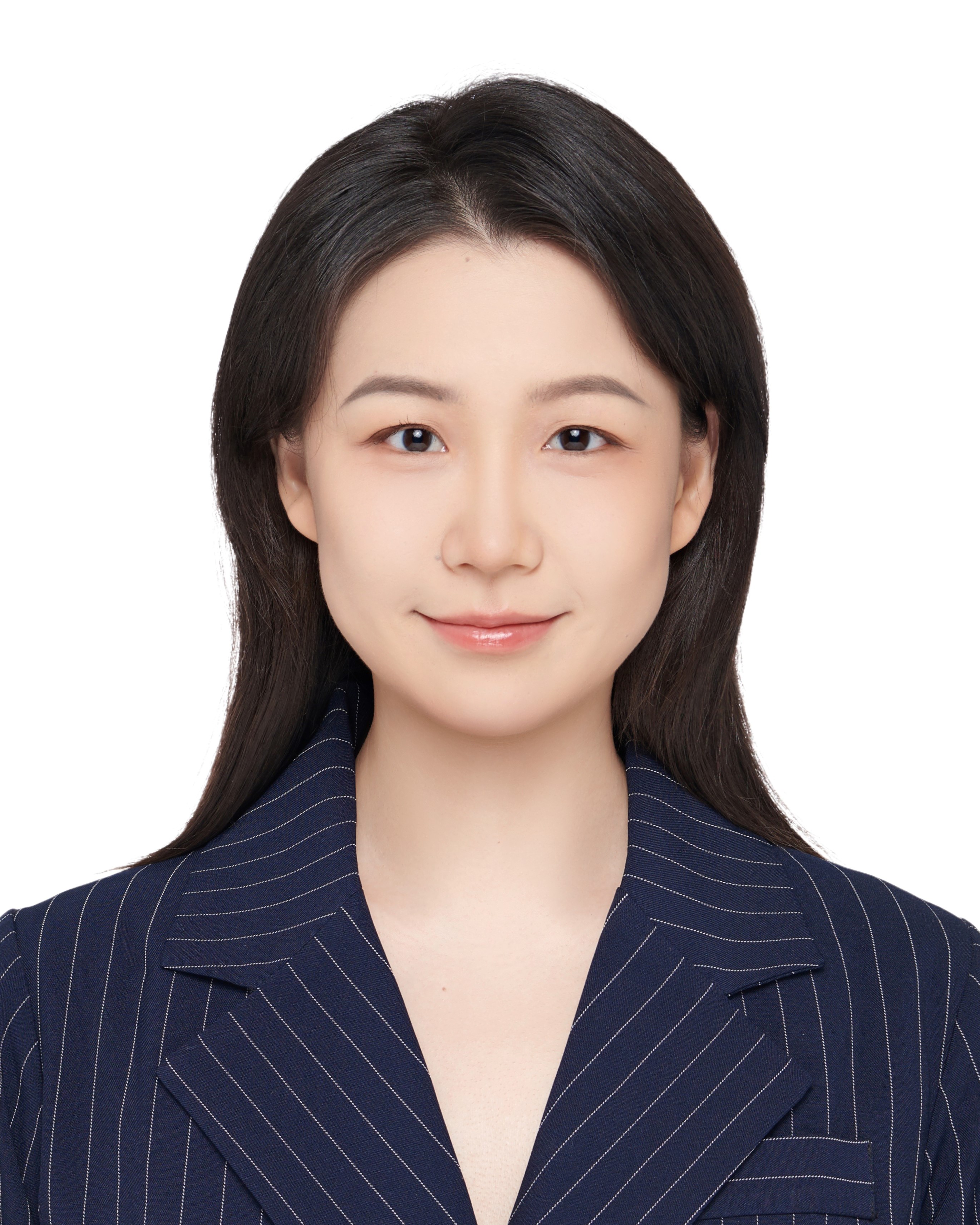}}]{Yueyuan LI} 
    received a Bachelor's degree in Electrical and Computer Engineering from the University of Michigan-Shanghai Jiao Tong University Joint Insitute, Shanghai, China, in 2020. She is working towards a Ph.D. degree in Control Science and Engineering from Shanghai Jiao Tong University.
    
    Her main fields of interest are the security of the autonomous driving system and driving decision-making. Her current research activities include driving decision-making models, driving simulation, and virtual-to-real model transferring.
\end{IEEEbiography}

\begin{IEEEbiography}[{\includegraphics[width=1in,height=1.25in,clip,keepaspectratio]{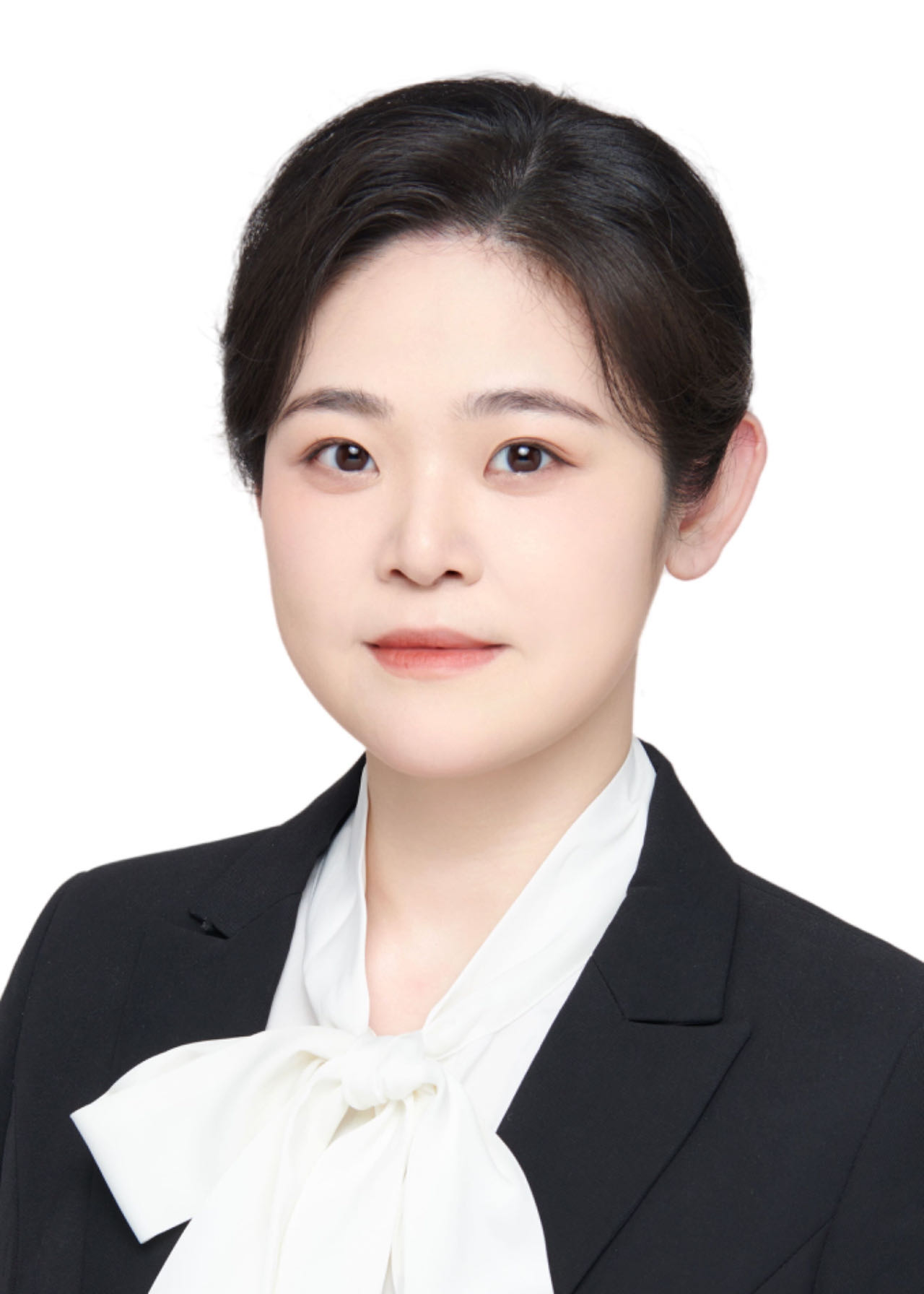}}]{Songan Zhang}
    received B.S. and M.S. degrees in automotive engineering from Tsinghua University in 2013 and 2016, respectively. Then, she went to the University of Michigan, Ann Arbor, and got a Ph.D. in mechanical engineering in 2021. After graduation, she worked at Ford Motor Company in the Robotics Research Team as a research scientist. Presently, she is an assistant professor at the Global Institute of Future Technology (GIFT) in Shanghai Jiao Tong University. Her research interests include accelerated evaluation of autonomous vehicles, model-based reinforcement learning, and meta-reinforcement learning for autonomous vehicle decision-making.    
\end{IEEEbiography}

\begin{IEEEbiography}[{\includegraphics[width=1in,height=1.25in,clip,keepaspectratio]{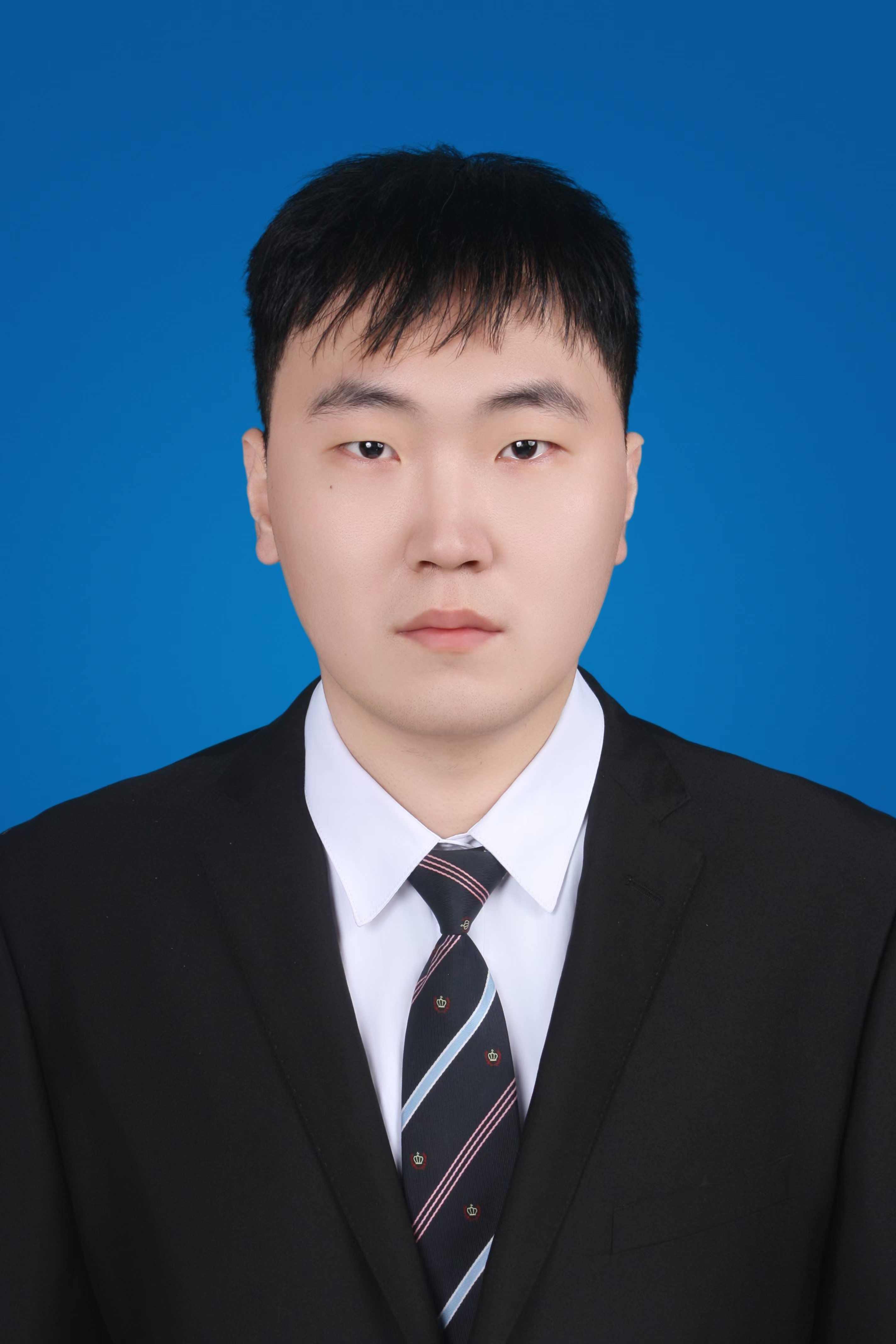}}]{Siyuan Chen}
    received a Bachelor's degree in engineering from Shanghai Jiao Tong University, Shanghai, China, in 2023. He is working towards a Master's degree in Control Science and Engineering from Shanghai Jiao Tong University. His main research interests are planning and control, V2X system for autonomous vehicles.
\end{IEEEbiography}

\begin{IEEEbiography}[{\includegraphics[width=1in,height=1.25in,clip,keepaspectratio]{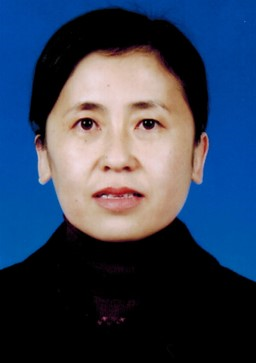}}]{Chunxiang WANG}
    received a Ph.D. degree in Mechanical Engineering from Harbin Institute of Technology, China, in 1999. She is currently an associate professor in the Department of Automation at Shanghai Jiao Tong University, Shanghai, China. 
    
    She has been working in the field of intelligent vehicles for more than ten years and has participated in several related research projects, such as European CyberC3 project, ITER transfer cask project, etc. Her research interests include autonomous driving, assistant driving, and mobile robots. 
\end{IEEEbiography}

\begin{IEEEbiography}[{\includegraphics[width=1in,height=1.25in,clip,keepaspectratio]{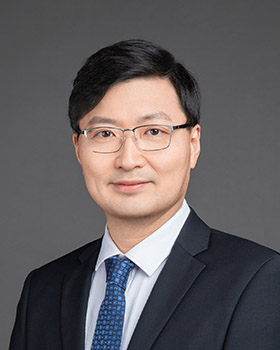}}]{Ming YANG}
    received his Master’s and Ph.D. degrees from Tsinghua University, Beijing, China, in 1999 and 2003, respectively. Presently, he holds the position of Distinguished Professor at Shanghai Jiao Tong University, also serving as the Director of the Innovation Center of Intelligent Connected Vehicles. Dr. Yang has been engaged in the research of intelligent vehicles for more than 25 years.
\end{IEEEbiography}

\clearpage
\appendices
\section{Implementation detials}
\label{appendix: state}


For the implementation of the information fusion transformer, we employ a multi-layer perceptron model for vector-based input to encode it into a feature token. For image-based input, we use a convolution neural network (CNN) with residual blocks to extract 2D features, which are later flattened to match the dimensions of other tokens. 
Tokens from different views are then fused in the transformer encoder, and their values are selected as outputs. Finally, an MLP-based network is used to map the concatenation of the values to the target output, namely the action for actor-net or the estimated value for value-net.
Moreover, we use an auto-encoder structure to train the encoder in self-supervision. The CNN encoder extracts the features from the original image, and a deconvolutional neural network is used to reconstruct the image from the features. The pre-training data is collected with a naive agent trained initially without image input. The loss function of the auto-encoder using a mean square error can be expressed as:
\begin{equation}
\begin{aligned}
    L_\text{AE} = \| \text{Decoder}(\text{Encoder}(I_{BEV}))-I_{BEV} \|_2^2.
\end{aligned}
\end{equation}

The key parameters for the algorithm and simulation are listed in Table \ref{tab: param1}. 
Besides, the training process was conducted on an NVIDIA GeForce RTX 3090 GPU and AMD EPYC 7542 CPU, and evaluation was performed on an NVIDIA GeForce RTX 3060 GPU and Intel i7-11800H CPU.

\section{Reward function}
\label{appendix: reward}
A basic reward design involves assigning a positive reward $r_{succ}$ when the goal is completed and a negative penalty $r_{fail}$ when the interaction terminates in failure. Based on this, we design several additional step rewards to encourage the agent to get close to the success state.

\subsubsection{Intersection-of-union(IOU) reward}
The IOU area of the vehicle bounding box and the target parking spot is selected as guidance. We denote $\text{IOU}(s_t)$ as the IOU area of vehicle at time step $t$, and the IOU reward is:
\begin{equation}
    r_{IOU}(t) = \text{max} \left(\text{IOU}(s_t) - \text{max}_{i \in (0,t-1)} \ \text{IOU}(s_i) , 0 \right).
\end{equation}
This reward focuses on the increase of the IOU area compared to the largest value reached before and does not penalize it when the vehicle attempts to move out of the parking spot.

\subsubsection{Distance reward}
We assign a positive reward to the agent when the vehicle is getting closer to parking spots:
\begin{equation}
\begin{aligned}
r_{dist}(t) = -\frac{\left( D(t) - D(0) \right)}{\text{max}\left(D(0), D_{min} \right)}.
\end{aligned}
\end{equation}
$D(t)$ is the distance from the vehicle to the target position at time step $t$. $D_{min}$ is a hyperparameter to avoid large rewards.

\subsubsection{Time consumption penalty}
For each interaction, the agent receives a small negative penalty as the cost of interaction, and this penalty increases over time. We bound the penalized value with $tanh$:
\begin{equation}
r_{time}(t)= -tanh(t / (10\cdot T_{max})).
\end{equation}
Here $T_{max}$ is the maximum interaction times in an episode.

\subsubsection{Overall reward function}
The above basic rewards and step rewards are linearly combined to form the final reward:
\begin{equation}
    r = w_1 r_{succ} + w_2 r_{fail} + w_3 r_{IOU} +w_4 r_{dist} +w_5 r_{time}.
\end{equation}
In practice, we set $r_{succ}=5,\ r_{fail}=-5$, $w_1=w_2=w_3=1$, $w_4=0.5$ and $w_5=0.1$.

\begin{table}[t]
  \vspace{-5pt}
   \caption{Parameters in simulation and algorithm}
  \centering
  \begin{tabular}{ c | c c }
    \hline
    \textbf{Parameter} & \textbf{Description} & \textbf{Value}  \\
    \hline
    $lr_{actor}$ & Learning rate for the actor-network & 5e-6  \\
    $lr_{critic}$ & Learning rate for the critic-network & 2.5e-5  \\
     $|D|$ & Replay buffer size & 8192 \\
     $\gamma$ & Discount factor for rewards & 0.98  \\
    $\epsilon$ & Clipping parameter in PPO & 0.2  \\
    $d_{rs}$ & Threshold distance for the RS policy & 10.0  \\
    $T_{max}$ & Maximum interaction times in a episode & 200 \\
    \hline
    $n_{MLP}$ & Number of layers in the input MLP  & 2  \\
    $n_{CNN}$ & Number of layers in the input CNN & 2  \\
    $d_{token}$ & The size of embedded tokens & 128  \\
    $d_{hidd}$ & The size of hidden layers  & 128  \\
    $n_{attn}$ & Number of attention layers & 1  \\
    $n_{head}$ & Parameters of multi-head mechanism & 8  \\
    $n_{out}$ & Number of layers in the output MLP  & 2  \\
    \hline
    $\Delta t$ & Simulated time for an interaction step & 0.5 s\\
    $\Delta \omega$ & Simulated lidar angular resolution & $\pi/60$ rad  \\
    $dim(l_t)$ & The dimension of obstacle distance vector & $120$ \\
    $dim(f_{am})$ & The dimension of action mask & $42$ \\
    $R_{lidar}$ & Simulated maximum lidar range & 10 m  \\
    $H_{img}(W_{img})$ & BEV image height (width) & 64 px  \\
    $L$ & Vehicle length & 4.69 m  \\
    $W$ & Vehicle width & 1.94 m  \\
    $v_{max}$ & Maximum velocity & 2.5 m/s  \\
    $\delta_{max}$ & Maximum steering angle & 0.75 rad  \\
    \hline
  \end{tabular}
  \vspace{-10pt}
  \label{tab: param1}
\end{table}

\section{Real-world experiment setups}
\label{appendix: real-world}

\begin{figure}[tb]
  \centering
  \includegraphics[width=0.8\linewidth]{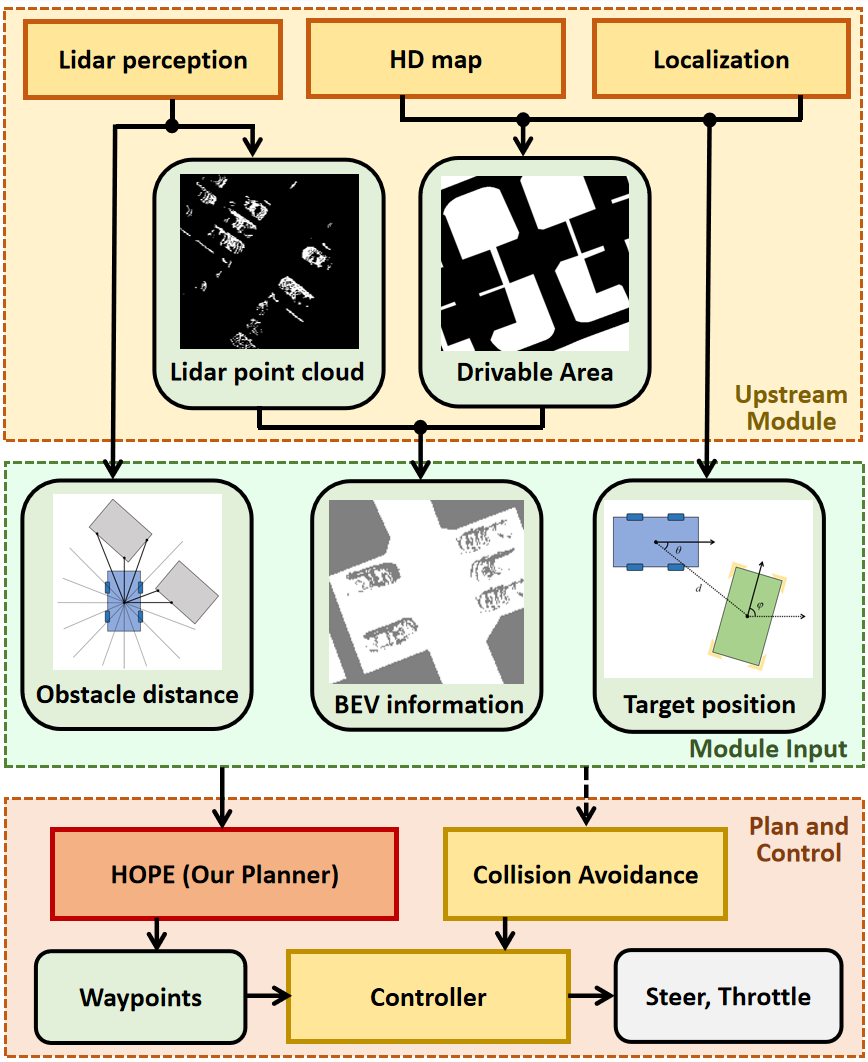}
  \caption{The overall structure of the parking system, where our proposed planning module is in red.}
  \label{fig: system}
\end{figure}

For sensors, we used Hesai Pandar40, RoboSense RS-LiDAR-32 
 and Bpearl LiDAR to support the vehicle's localization and perception modules and no GPS was included.
Figure \ref{fig: system} illustrates the framework of the parking system and how our method operates in real-world scenarios. The input to the planning module, where our method is applied, consists of a BEV grid map, obstacle distance, and the position of the parking spot. The target parking spot is selected by the driver and its precise location is obtained from a high-definition map, then transformed into the ego coordinate system through the localization system. Obstacle distance information is derived from the LiDAR point cloud, while the BEV grid map is generated by augmenting the drivable area from the high-definition map with real-time detected obstacles from the LiDAR. To mitigate the impact of upstream perception errors, we also add virtual occupancy grids in other parking spaces in the BEV grid map based on high-definition map information, ensuring the vehicle does not drive into these areas during maneuvering. 
These inputs are then converted into the format required by the network, which outputs a one-step planned path. Finally, a Quadratic Programming (QP) and Model Predictive Control (MPC) approach is used for path following to obtain the control commands from the waypoints. 
Besides, we also utilize a collision avoidance module to give way to dynamic obstacles in practice. 
The vehicle only follows a path if it has been successfully planned as collision-free and is connected to the target parking position. All modules and our algorithm are deployed on an Intel i7-1165G7 2.80GHz CPU.

\section{Experiments on the Extended Reeds-Shepp Curves}
\label{appendix: scs}
\begin{figure}[hb]
    \vspace{-10pt}
  \centering
  \includegraphics[width=0.9\linewidth]{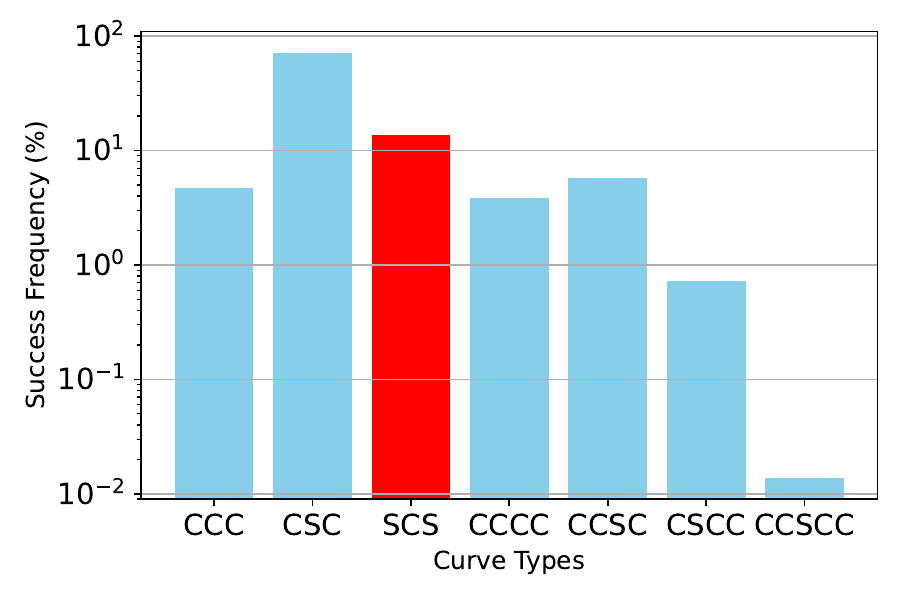}
    \vspace{-10pt}
  \caption{The success frequency of different RS curve types}
  \label{fig: rs-type}
\end{figure}

We conducted experiments to evaluate the impact of different Reeds-Shepp (RS) curve types on the success rate of parking maneuvers. The results are summarized in Figure \ref{fig: rs-type}, where the horizontal axis represents different RS curve types, and the vertical axis represents the frequency of usage of each curve type in successful planning cases for all vertical parking scenarios. The results show that SCS curves have the second-highest frequency of 13.8\%,  demonstrating their effectiveness.

\color{black}
\section{Proof of Convergence for the Hybrid Policy}
The convergency of the update on the parameter $V_\psi$ and $Q_\phi$ can be proven by showing the update process of the Bellman expectation operator is a contraction mapping, which is independent of specific policy $\pi$. Thus the convergency stands for any given $\pi$ and so is the $\pi_h$. We here provide the proof of convergency of the value function $V_\psi$, and the Q function $Q_\phi$ can be obtained using a one-step iteration:
\begin{equation}
    Q^{\pi}(s, a) = \mathbb{E} \left[ R_{t+1} + \gamma V^{\pi}(S_{t+1}) \mid S_t = s, A_t = a \right]
\end{equation}

Firstly, the Bellman expectation equation for a policy $\pi$ is defined as:
\begin{equation}
V_\pi(s_t) = \mathbb{E}_{a_t \sim \pi(\cdot|s_t)}[r(s_t, a_t) + \gamma V_\pi(s_{t+1})]
\end{equation}

We define the Bellman expectation operator $B_\pi$ as:
\begin{equation}
(B_\pi V)(s_t) = \mathbb{E}_{a \sim \pi(\cdot|s)}[r(s_t, a) + \gamma V(s_{t+1})]
\end{equation}

To prove that the Bellman expectation operator $B_\pi$ is a contraction mapping, we need to show:
\begin{equation}
\|B_\pi V_1 - B_\pi V_2\| \leq \gamma \|V_1 - V_2\|
\end{equation}

For any two value functions $V_1$ and $V_2$, we have:
\begin{align*}
& |B_\pi V_1(s_t) - B_\pi V_2(s_t)| \\
& = \left| \mathbb{E}_{a \sim \pi(\cdot|s_t)} \left[ r(s_t, a) + \gamma \mathbb{E}_{s_{t+1} \sim E} [V_1(s_{t+1})] \right]  \right.\\
& - \left. \mathbb{E}_{a \sim \pi(\cdot|s)} \left[ r(s_t, a) + \gamma \mathbb{E}_{s_{t+1} \sim E} [V_2(s_{t+1})] \right] \right| \\
&= \gamma \left| \mathbb{E}_{a \sim \pi(\cdot|s_t)} \left[ \mathbb{E}_{s_{t+1} \sim E} [V_1(s_{t+1}) - V_2(s_{t+1})] \right] \right| \\
&= \gamma \left| \mathbb{E}_{a \sim \pi(\cdot|s_t), s_{t+1} \sim E} [V_1(s_{t+1}) - V_2(s_{t+1})]  \right| \\
&\leq \gamma \mathbb{E}_{a \sim \pi(\cdot|s_t), s_{t+1} \sim P(\cdot|s_t, a)} \left[ \left| V_1(s_{t+1}) - V_2(s_{t+1}) \right| \right]  \\
&\leq \gamma \sup_{s_{t+1}} |V_1(s_{t+1}) - V_2(s_{t+1})| \\
&= \gamma \|V_1 - V_2\|_\infty
\end{align*}

This shows that $B_\pi$ is a contraction mapping because $\|B_\pi V_1 - B_\pi V_2\|_\infty \leq \gamma \|V_1 - V_2\|_\infty$ for $0 \leq \gamma < 1$.

According to the Banach fixed-point theorem, a contraction mapping $B_\pi$ on a complete metric space $(\mathbb{R}^S, \|\cdot\|_\infty)$ has a unique fixed point $V_\pi^*$, i.e.,
\begin{equation}
B_\pi V_\pi^* = V_\pi^*
\end{equation}

Therefore, through the iterative update process:
\begin{equation}
V_{k+1} = B_\pi V_k
\end{equation}
the value function $V_k$ will converge.

The convergence proof above shows that the Bellman expectation operator $B_\pi$ is a contraction mapping and hence will converge to a fixed point for any given policy $\pi$. This means that the convergence of the value function updates is not dependent on the specific policy $\pi$. Therefore, even if we replace the policy $\pi$ with another policy $\pi_h$, the convergence properties still hold. We can use the following loss function to update the value function:
\begin{equation}
J_V(\psi) = \mathbb{E}_{s_t \sim D_{\pi_h}} \left[ \frac{1}{2} \left( V_\psi(s_t) - \mathbb{E}_{\tilde{a}_t \sim \pi_h} \left[ Q_\phi(s_t, \tilde{a}_t) \right] \right)^2 \right]
\end{equation}

We minimize this loss function to update the parameter $\psi$:
\begin{equation}
\psi_{k+1} = \psi_k - \alpha \nabla_\psi J_V(\psi_k)
\end{equation}

This loss function minimizes the mean squared error between $V_\psi(s_t)$ and $\mathbb{E}_{\tilde{a}_t \sim \pi_h} \left[ Q_\phi(s_t, \tilde{a}_t) \right]$. According to the Bellman expectation equation:
\begin{equation}
V_\pi(s_t) = \mathbb{E}_{\tilde{a}_t \sim \pi_h} \left[ Q_\phi(s_t, \tilde{a}_t) \right]
\end{equation}
minimizing $J_V(\psi)$ is equivalent to approximating the Bellman expectation equation.

\end{document}